\documentclass{article}
\PassOptionsToPackage{numbers, compress}{natbib}
\usepackage[preprint]{neurips_2020}
\usepackage{hyperref}
\usepackage{url}
\usepackage{amsmath, amssymb, amsthm, cleveref}
\usepackage{wrapfig}
\crefname{equation}{}{}
\Crefname{equation}{}{}

\newcommand{\comment}[1]{}
\usepackage[T1]{fontenc}    
\usepackage{hyperref}       
\usepackage{url}            
\usepackage{booktabs}       
\usepackage{amsfonts}       
\usepackage{nicefrac}       
\usepackage{microtype}      
\usepackage{algorithm}
\usepackage{algorithmic}

\usepackage[utf8]{inputenc}
\usepackage{mathtools}
\usepackage{subcaption}
\usepackage{graphicx}
\usepackage{caption}
\usepackage{bbm}
\usepackage[shortlabels]{enumitem}
\usepackage{array}
\usepackage{arydshln}

\usepackage{multirow}
\crefname{equation}{}{}
\Crefname{equation}{}{}

\usepackage{url}            
\usepackage{booktabs}       
\usepackage{amsfonts}       
\usepackage{nicefrac}       
\usepackage{xcolor}
\usepackage{microtype}      
\usepackage{amssymb,amsmath, amsthm}
\usepackage[utf8]{inputenc}
\usepackage{mathtools}
\usepackage{subcaption}
\usepackage{tikz}
\usetikzlibrary{calc}
\usepackage{graphicx}
\usepackage{caption}
\usepackage{bbm}
\usepackage{bm}

\usepackage[shortlabels]{enumitem}
\usepackage{array}
\usepackage{arydshln}
\newtheorem{theorem}{Theorem}[section]
\newtheorem{lemma}{Lemma}[section]
\newtheorem{assumption}{Assumption}[section]

\DeclareMathOperator*{\leqt}{\leq}

\usepackage[colorinlistoftodos,prependcaption]{todonotes}

\definecolor{gr}{rgb}{0.25, 0.25, 0.25}

\newcommand{\q}{q}

\newcommand{\ec}{K}

\newcommand{\ic}{k}

\newcommand{\pkt}{\textsc{PerFed-CKT }}
\newcommand{\pktc}{\textsc{PerFed-CKT}}

\newcommand\tikzmark[1]{\tikz[remember picture,overlay] \coordinate (#1);}

\newcommand{\gf}{F}

\newcommand{\lff}{f}
\newcommand{\lf}{F_\ic}
\newcommand{\gl}{\mathcal{L}}

\newcommand{\sigl}{s}
\newcommand{\siglc}{\mathbf{s}}
\newcommand{\siglb}{\overline{\mathbf{s}}}
\newcommand{\cent}{\mathbf{c}}
\newcommand{\nc}{c}
\newcommand{\sw}{\alpha}
\newcommand{\dd}{\mathcal{D}}
\newcommand{\ddh}{\widehat{\mathcal{D}}}

\newcommand\inner[2]{\left\langle #1, #2 \right\rangle}

\newcommand{\gb}{\mathbf{g}}
\newcommand{\gc}{\Phi}
\newcommand{\wb}{\mathbf{w}}
\newcommand{\xb}{\mathbf{x}}
\newcommand{\wbt}{\widehat{\mathbf{w}}}
\newcommand{\wbl}{\widetilde{\mathbf{w}}}
\newcommand{\zb}{\mathbf{z}}
\newcommand{\yb}{\mathbf{y}}
\newcommand{\xxb}{\mathbf{X}}

\newcommand{\pb}{\mathbf{P}}

\newcommand{\ib}{\mathbf{I}}
\newcommand{\eb}{\mathbf{e}}

\newcommand{\bdat}{\mathcal{B}_\ic}

\newcommand{\ldat}{m_\ic}

\newcommand{\rdat}{p_k}

\newcommand{\ti}{(t)}
\newcommand{\tip}{(t+1)}

\newcommand{\midat}{\xi_{\ic}^{\ti}}

\newcommand{\sgrad}{\mathbf{g}_\ic}
\newcommand{\hgrad}{h_\ic}
\newcommand{\grad}{G}

\newcommand{\lr}{\eta_t}

\newcommand{\expt}{\mathbb{E}}

\DeclareMathOperator*{\argmin}{arg\,min} 

\title{Personalized Federated Learning for Heterogeneous Clients with Clustered Knowledge Transfer} 
\author{
  Yae Jee Cho \\
  ECE Department\\
  Carnegie Mellon University\\
  Pittsburgh, PA 15213 \\
  \texttt{yaejeec@andrew.cmu.edu} 
  \And
  Jianyu Wang \\
  ECE Department\\
  Carnegie Mellon University\\
  Pittsburgh, PA 15213 \\
  \texttt{jianyuw1@andrew.cmu.edu} 
  \And
  Tarun Chiruvolu\\
  School of Computer Science\\
  Carnegie Mellon University\\
  Pittsburgh, PA 15213 \\
  \texttt{tchiruvo@andrew.cmu.edu}
  \And
  Gauri Joshi \\
  ECE Department\\
  Carnegie Mellon University\\
  Pittsburgh, PA 15213 \\
  \texttt{gaurij@andrew.cmu.edu} 
}

\begin{document}

\maketitle

\begin{abstract}
Personalized federated learning (FL) aims to train model(s) that can perform well for individual clients that are highly data and system heterogeneous. Most work in personalized FL, however, assumes using the same model architecture at all clients and increases the communication cost by sending/receiving models. This may not be feasible for realistic scenarios of FL. In practice, clients have highly heterogeneous system-capabilities and limited communication resources. In our work, we propose a personalized FL framework, \pktc, where clients can use heterogeneous model architectures and do not directly communicate their model parameters. \pkt uses \emph{clustered co-distillation}, where clients use logits to transfer their knowledge to other clients that have similar data-distributions. We theoretically show the convergence and generalization properties of \pkt and empirically show that \pkt achieves high test accuracy with several orders of magnitude lower communication cost compared to the state-of-the-art personalized FL schemes. 

\end{abstract}

\section{Introduction}
The emerging paradigm of federated learning (FL) \cite{mcmahan2017communication, kairouz2019advances, bonawitz2019towards,wang2021field} enabled the use of data collected by thousands of resource-constrained clients to train machine learning models without having to transfer the data to the cloud. Most recent work \cite{yu2018parallel, stich2018local, wang2018cooperative, yjc2020csfl} is focused on algorithms for training a single global model with edge-clients via FL. 
However, due to the inherently high data-heterogeneity across clients \cite{reddi2020adaptive, haddadpour2019convergence, khaled2020tighter, stich2019error, woodworth2020local, koloskova2020unified, huo2020faster, zhang2020fedpd, pathak2020fedsplit, malinovsky2020local, sahu2019federated}, a single model that is trained to perform best in expectation for the sum of all participating clients' loss functions may not work well for each client \cite{liang2020thinklocal, ghosh2020cfl, ziyi2020tpfl}. For example, if a single global model was trained for next-word prediction with all available clients and the input was ``I was born in'', the model will most likely give bad results for the individual clients.

The limited generalization properties of conventionally trained FL models to clients with scarce local data calls for the design of FL algorithms to train personalized models that can perform well for individual clients. Several works have investigated personalized FL~\cite{fallah2020personalized, zha2021fedfomo, li2021ditto, mansour2020approaches}, including applying meta-learning~\cite{fallah2020personalized}, training separate models on each client with weighted aggregation of other clients' models~\cite{zha2021fedfomo}, using the global objective as a regularizer for training individual models at each client~\cite{li2021ditto}, or using model/data-interpolation with clustering for personalization~\cite{mansour2020approaches}. However, the aforementioned work does not consider two critical factors of FL: i) the computation and memory capabilities can be heterogeneous across clients and ii) the cost of communicating high-dimensional models with the server can be prohibitively high~\cite{sha2017largenn,chao2020GKT}. Most work in personalized FL assumes a homogeneous model architecture across clients, and frequent communication of the model parameters. 

In this work, we propose training personalized models with clustered co-distillation. Co-distillation~\cite{shag2020codist, bist2020distdis,lin2020ensemble} is an approach to perform distributed training across different clients with reduced communication cost by only exchanging models' predictions on a common unlabeled dataset instead of the model parameters. This method adds a regularization term to the local loss of each client to penalize the client's prediction from being significantly different from the predictions of other clients. In the conventional co-distillation, the regularizing term for each client is the average of all other participating clients' predictions. However in FL, clients' data can be highly heterogeneous. Thus, forcing each client to follow the average prediction of all clients can exacerbate its generalization by learning irrelevant knowledge from clients that have significantly different data distributions~\cite{anil2020codistill}. Hence, we propose a novel \emph{clustered co-distillation} framework \pktc, where each client uses the average prediction of only the clients that have similar data distributions. This way, we prevent each client from assimilating irrelevant knowledge from unrelated clients. 

In short, \pkt largely improves on current personalized FL strategies in the following ways: 
\begin{itemize}
    \item Allows model heterogeneity across clients where the architecture and size of the model for local training can vary across clients.
    \item Dramatically reduces the communication cost by transferring logits instead of model parameters between the clients and the server.
    \item Improves generalization performance for data-scarce clients, while preventing learning from unrelated clients by clustered knowledge transfer.
\end{itemize}
We also present theoretical analysis of \pkt with its convergence guarantees and generalization performance. The generalization results show that clustering indeed helps in terms of improvement of the generalization properties of individual clients. Our experiments demonstrate that for both model-homogeneous and model-heterogeneous environments, \pkt can achieve high test accuracy with several orders of magnitude less communication.

\section{Background and Related Work}

\paragraph{Personalized FL.} 
In personalized FL, the goal is to train a single or several model(s) that can generalize well to each client's test dataset. In~\cite{fallah2020personalized}, using meta-learning for training a global model that better represents each client's data was proposed. A similar line of work using the moreau envelope as a regularizer was proposed in \cite{dinh2020pfedme}. Work in \cite{zha2021fedfomo} proposed to find the optimal weighted combination of models from clients so that each client gets a model that better represents its target data distribution. The authors in \cite{mansour2020approaches} propose general approaches that can be applied to vanilla FL for personalization, including client clustering and data/model interpolation. 

The aforementioned work however, all requires model homogeneity and direct communication of model parameters across clients/server. Although \cite{li2019fedmd} does consider communication cost in personalization by using distillation, it does not provide any theoretical guarantees and does not consider the high data heterogeneity across clients that can tamper with the personalization performance. Moreover they require the presence of a \textit{large labeled} public dataset, which is realistically an expensive resource to have access to. Our work investigates a novel personalized FL framework that allows model heterogeneity, improves communication-efficiency, and utilizes data heterogeneity with clustering while using a \textit{small unlabeled} public dataset. 

\begin{figure*}[!h]
\centering
\begin{subfigure}{0.24\textwidth}
\centering
\includegraphics[width=1\textwidth,height=0.85\textwidth]{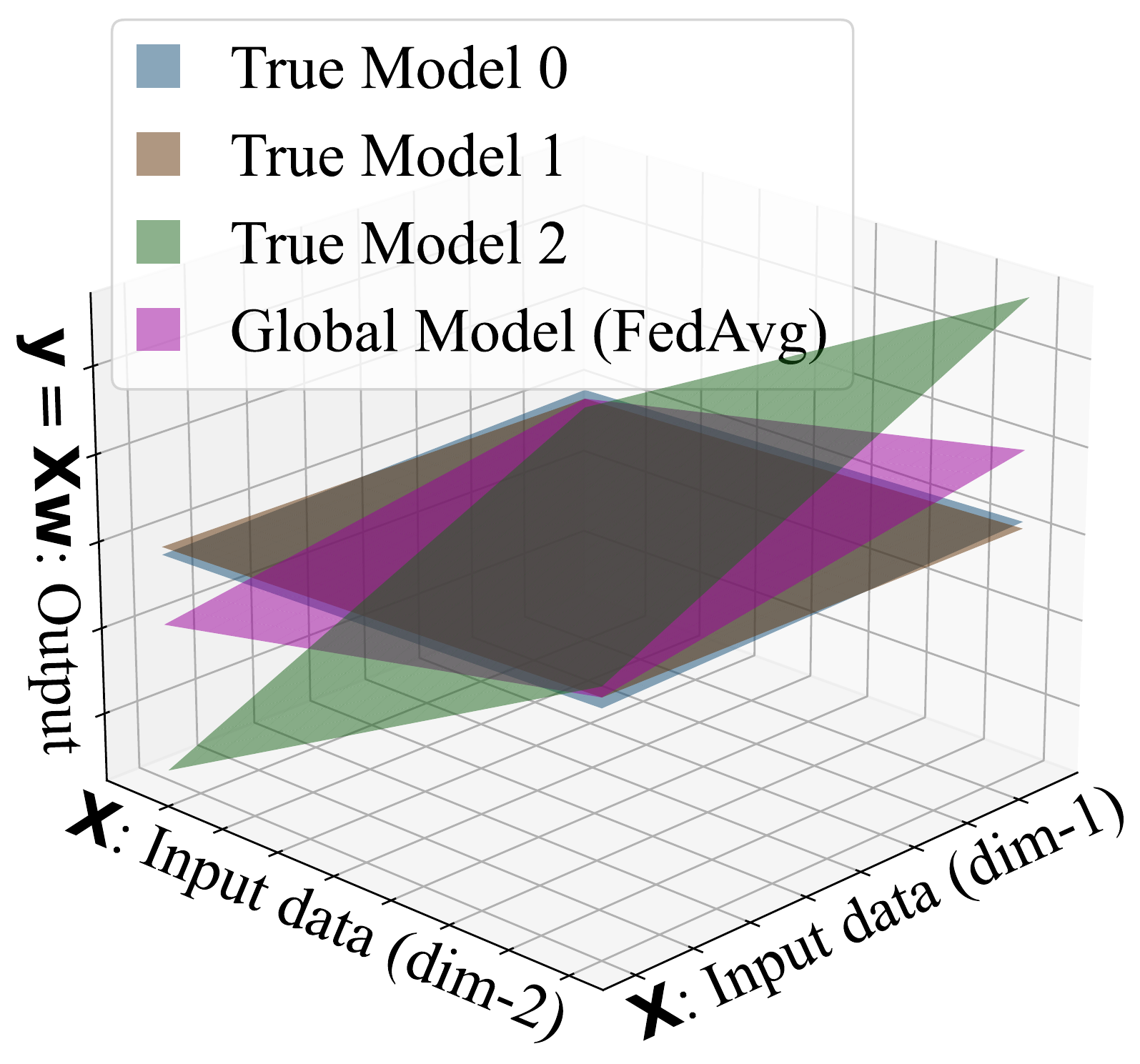} \label{fig:0-1} \vspace{-1.6em}
\caption{}
\end{subfigure}
\begin{subfigure}{0.24\textwidth}
\centering
\includegraphics[width=1\textwidth]{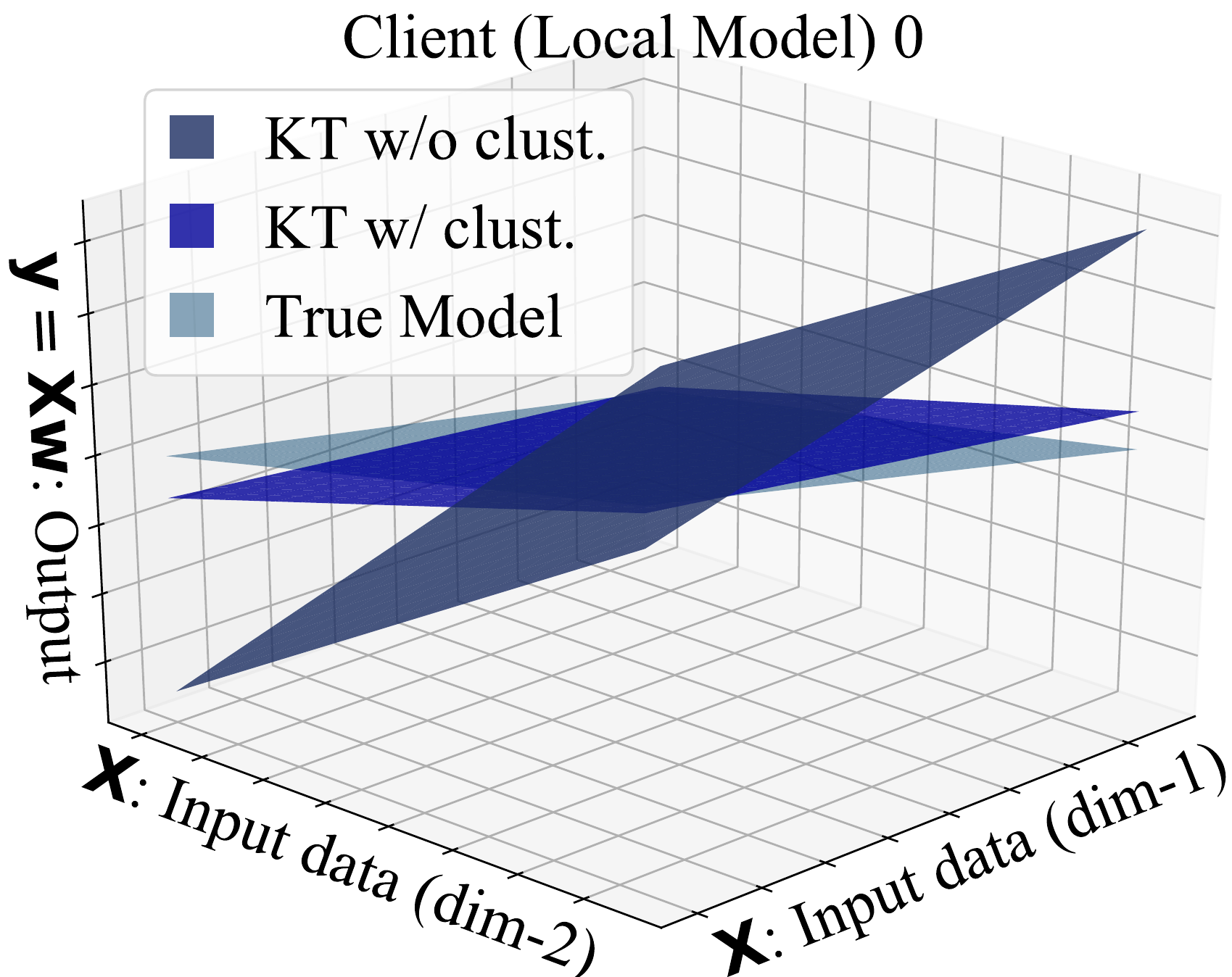} 
\caption{}
\end{subfigure}
\begin{subfigure}{0.24\textwidth}
\centering
\includegraphics[width=1\textwidth]{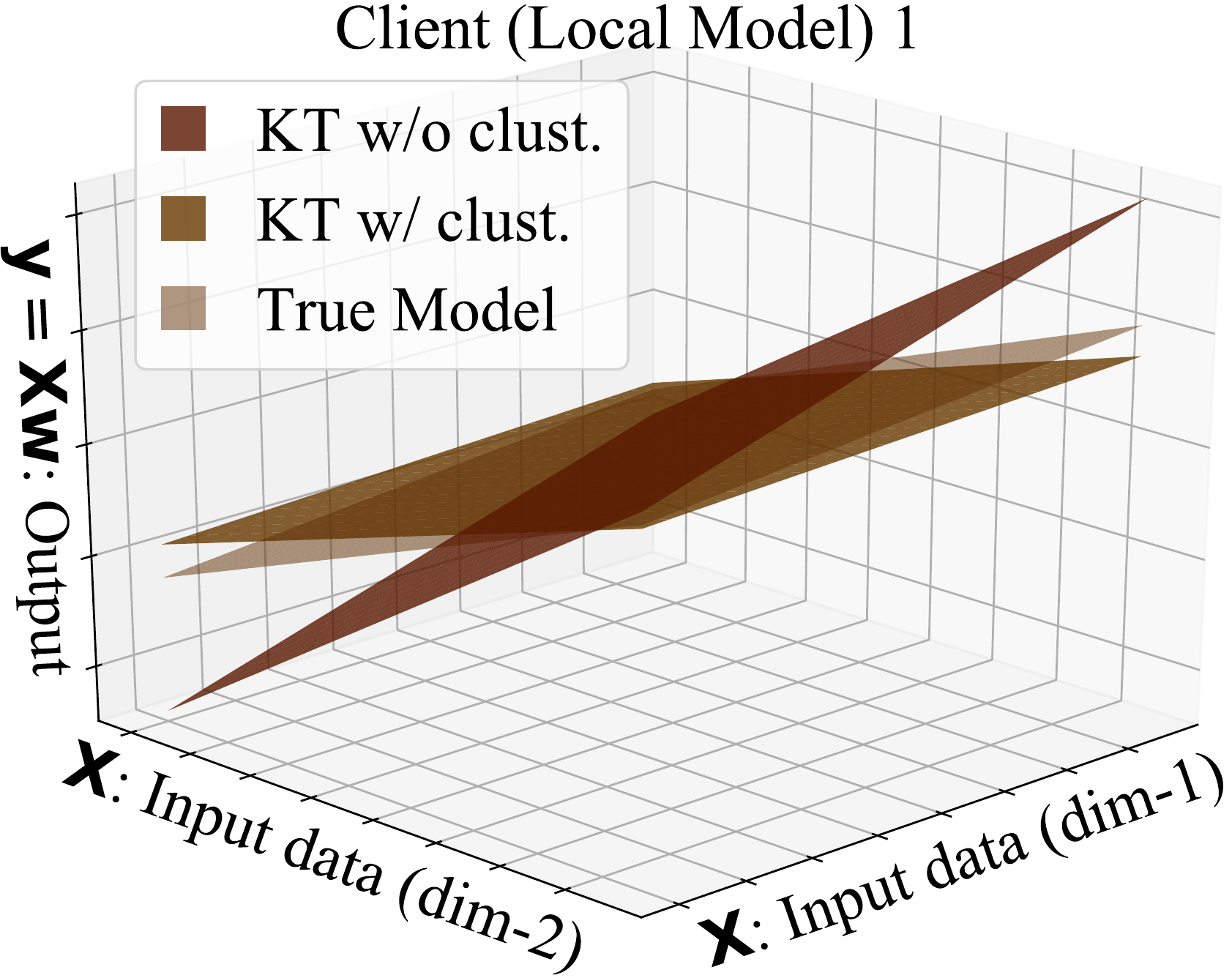} 
\caption{}
\end{subfigure}
\begin{subfigure}{0.24\textwidth}
\centering
\includegraphics[width=1\textwidth]{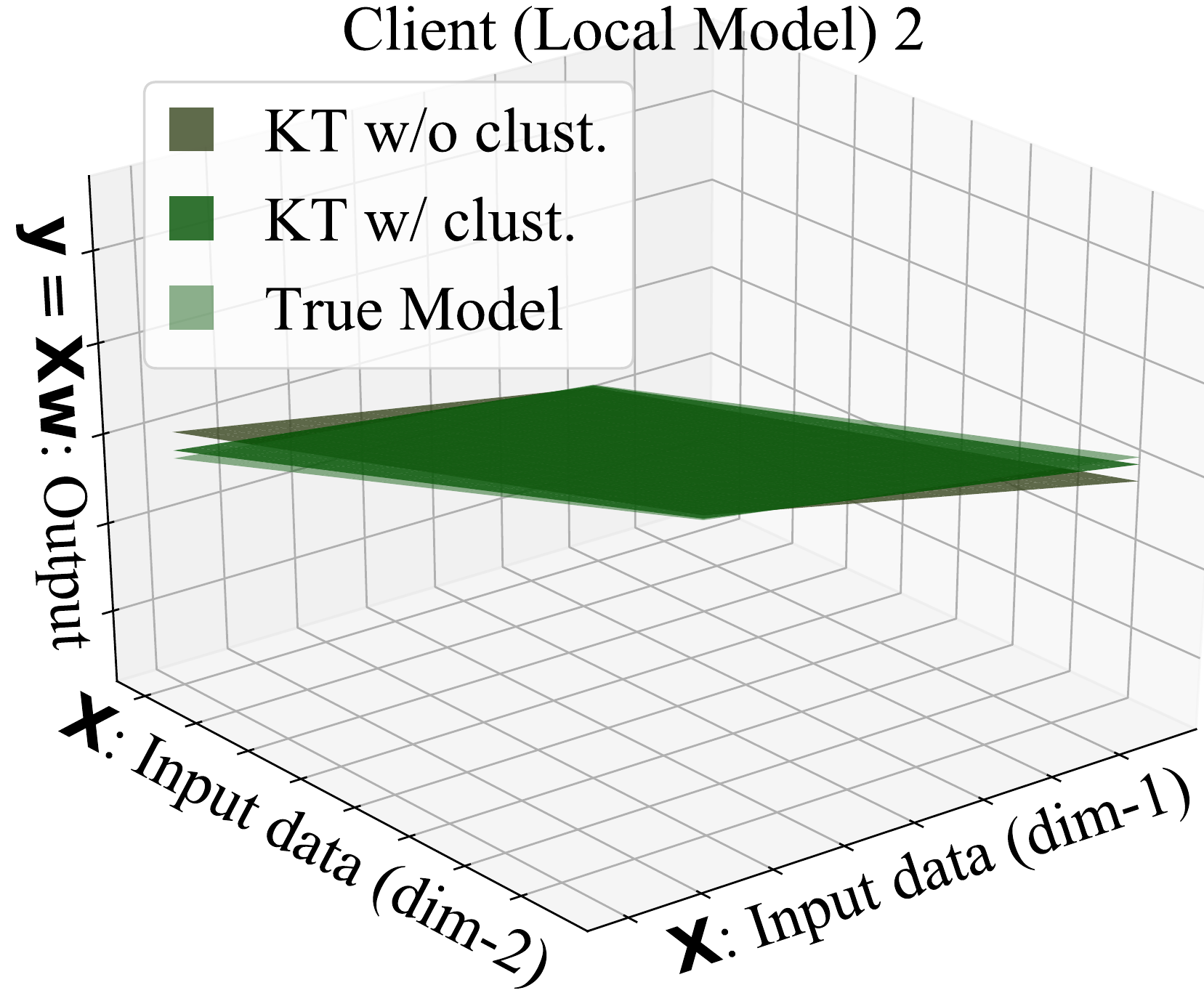} \vspace{-1.3em}
\caption{}
\end{subfigure} 
\caption{Toy example with linear regression for given input data $\mathbf{X}\in\mathbb{R}^2$ and local (true) models $\mathbf{w}\in\mathbb{R}^{2\times1}$ for three clients indexed by 0-2; (a): the true model for each client and the global model from FedAvg. The resulting global model does not match well with clients' true model; (b)-(d): the model for each client resulting from \pkt with and without clustering (simple average of logits). \pkt with clustering yields the model closest to the true model for all clients.
}
  \label{fig:lrtoy} 
\vspace{-1em}
\end{figure*}

\paragraph{Knowledge Transfer.} Knowledge distillation (KD)~\cite{hinton2015distillknow} is prominently used as a method of knowledge transfer from a pre-trained larger model to a smaller model~\cite{ma2020distillhet,ita2021distillsemisup,sun2021distillpriv, qinli2021distillone,zhouyan2021distilloneshot,sangho2020distilledge, eun2018dist,lan2018distillfly,lyu2020flprivacy}. Extending from this conventional KD, co-distillation~\cite{anil2020codistill,zhan2018dml,zhan2021adcodis} transfers knowledge across multiple models that are being trained concurrently. Specifically, each model is trained with the supervised loss with an additional regularizer term that encourages the model to yield similar outputs to the outputs of the other models that are also being trained. 

Using co-distillation for improved generalization in distributed training has been recently proposed in \cite{shag2020codist, bist2020distdis,lin2020ensemble}. Authors of \cite{shag2020codist} have shown empirically that co-distillation indeed improves generalization for distributed learning but often results in over-regularization, where the trained model's performance drops due to overfitting to the regularization term. In \cite{bist2020distdis}, co-distillation was suggested for communication-efficient distributed training, but not in the personalized FL context where data can be highly heterogeneous across nodes and presented limited experiments on a handful of nodes with homogeneous data distributions across the nodes. 

Applying co-distillation for personalized FL presents a unique challenge in that each client's data distribution can be significantly different from other clients' distributions. Using standard co-distillation with the entire clients' models can actually worsen clients' test performance due to learning irrelevant information from other clients with different data distributions. We show in our work that this is indeed the case and show that using clustering to find the clients that have similar data-distributions with each other and then performing co-distillation within the clusters improves the personalized model's performance significantly for each client. 


\section{Proposed Personalized FL Framework: \pkt}

\subsection{Problem Formulation}
 \label{sec:pkt}
 
Consider a cross-device FL setup where a large number of $\ec$ clients (edge-devices) are connected to a central server. We consider a $N$-class classification task where each client $\ic\in[\ec]$ has its local training dataset $\bdat$ with $|\bdat|=\ldat$ data samples. We denote $\rdat=\ldat/{\sum_{\ic=1}^{\ec}\ldat}$ as the fraction of data for client $\ic$. Each data sample $\xi$ is a pair $(\xb,y)$ where $\xb\in\mathbb{R}^d$ is the input and $y\in[1,N]$ is the label. The dataset $\bdat$ is drawn from the local data distribution $\dd_\ic$, where we denote the empirical data distribution of $\bdat$ as $\ddh_\ic$. In the standard FL~\cite{mcmahan2017communication}, clients aim to collaboratively find the model parameter vector $\wb\in\mathbb{R}^{n}$ that maps the input $\xb$ to label $y$, such that $\wb$ minimizes the empirical risk $\gf(\wb) = \sum_{\ic=1}^\ec \rdat \lf(\wb)$. The function $\lf(\wb)$ is the local objective of client $\ic$, defined as $\lf(\wb)=\frac{1}{|\bdat|}\sum_{\xi \in \bdat} f(\wb,\xi)$ with $f(\wb, \xi)$ being the composite loss function. 

Due to high data heterogeneity across clients, the optimal model parameters $\wb^*$ that minimize $\gf(\wb)$ can generalize badly to clients whose local objective $\lf(\wb)$ significantly differs from $\gf(\wb)$. Such clients may opt out of FL, and instead train their own models $\wb_\ic\in\mathbb{R}^{n_\ic}$ by minimizing their local objectives, where $\wb_\ic$ can be heterogeneous in dimension. This can work well for clients with a large number of training samples (i.e., large $\ldat$), since their empirical data distribution $\ddh_\ic$ becomes similar to $\dd_\ic$, ensuring good generalization. However, if clients have a small number of training samples, which is often the case~\cite{lin2021flpos}, the distributions $\dd_\ic$ and $\ddh_\ic$ can differ significantly, and therefore a model $\wb_\ic$ trained only using the local dataset $\bdat$ can generalize badly. Hence, although clients with small number of training samples are motivated to participate in FL, the clients may actually not benefit from FL due to the bad generalization properties coming from other clients with significantly different data distributions. We show that with our proposed \pktc, clients with small training samples still benefit from participating in FL, improving generalization by being clustered with clients with similar data distributions. 

\paragraph{Objective with Clustered Knowledge Transfer.} We use co-distillation across different clients in FL for personalization, where each client co-distills with only the other clients that have similar model outputs with its own model, namely, \textit{clustered knowledge transfer}. With clustered knowledge transfer, each client learns from clients that have similar data distributions to improve its generalization performance.
 Formally, we consider each client having access to its private dataset $\bdat$ and a public dataset $\mathcal{P}$, consisting of \textit{unlabeled} data. The public dataset $\mathcal{P}$ is used as a reference dataset for co-distillation across clients\footnote{Unlike the majority of existing distillation methods which require expensive labeled data, we show the feasibility of leveraging unlabeled datasets. Note that unlabeled data can be either achieved by existing datasets or a pre-trained generator (e.g., GAN).}. The classification models $\wb_\ic,~\ic\in[\ec]$ output soft-decisions (logits) over the pre-defined number of classes $N$, which is a probability vector over the $N$ classes. We refer to the soft-decision of model $\wb_\ic$ over any input data $\xb$ in either the private or public dataset as $\sigl(\wb_\ic, \xb):\mathbb{R}^{n_\ic}\times(\mathcal{B}_\ic\cup\mathcal{P})\rightarrow\Delta_N$, where $\Delta_N$ stands for the probability simplex over $N$. For notational simplicity, we define $\siglc_\ic\in\mathbb{R}^{|\mathcal{P}|\times N}$ as $\sigl(\wb_\ic,\xb)\in\mathbb{R}^{1\times N}$, $\xb\in\mathcal{P}$ stacked into rows for each $\xb$. We similarly define $\siglb_\ic=\sum_{i=1}^K  \sw_{k,i}\siglc_i$. 

The clients are connected via a central aggregating server. Each client seeks to find the model parameter $\wb_\ic$ that minimizes the empirical risk $\gc_\ic(\wb_\ic;\siglb_\ic)$, where $\gc_\ic(\wb_\ic;\siglb_\ic)$ is a sum of the empirical risk of its own local training data $\lf(\wb_\ic)$ and the regularization term as follows:
\begin{align}
\gc_\ic(\wb_\ic;\siglb_\ic)=\lf(\wb_\ic)+\underbrace{\frac{\lambda}{|\mathcal{P}|}\sum_{\xb\in\mathcal{P}}\|\siglb_{\ic}(\xb)-\sigl(\wb_\ic,\xb)\|_2^2}_{\text{regularization term}} \label{eqn:main_objective}
\end{align}
The term $\siglb_\ic(\xb)=\sum_{i=1}^K \sw_{k,i}\sigl(\wb_i,\xb)$ denotes the weighted average of the logits from all clients for an arbitrary set of weights for client $k$, i.e., $\{\sw_{k,i}\}_{i\in[\ec]}$ such that $\sum_{i=1}^K\sw_{k,i}=1,~\forall k\in[K]$.  The term $\lambda$ modulates the weight of the regularization term. The weight $\sw_{k,i}$ for each client $i,~i\in[K]$ with respect to client $\ic$ results from clustering the logits by the $\ell_2$-norm distance so that clients with similar logits will have higher weights for each others' logits. The aggregated logit information with weights, $\siglb_\ic$, is calculated and sent by the server to the clients. Details of how the weights $\sw_{k,i},{i\in[\ec]}$ for each client $\ic$ are calculated and how the logits are communicated are elaborated in more detail in the subsequent Algorithm subsection. Before going into details of the algorithm we first give more intuition on the formulation of the regularization term in the next paragraph. 

\paragraph{Regularization Term.} Without the regularization term in \cref{eqn:main_objective}, minimizing $\gc_\ic(\wb_\ic;\siglb_\ic)$ with regards to $\wb_\ic$ is analogous to locally training in isolation for minimizing the local objective function $\lf(\wb_\ic)$ for client $\ic$. If we have the regularization term with $\sw_{k,i}=1/K,~\forall~k,~i\in[\ec]$, co-distillation is implemented without clustering, using all of the clients' knowledge. We show in the next toy example and in a generalization bound derived for ensemble models in personalization in \Cref{app:genbound} of why setting the values $\sw_{k,i},k,i\in[K]$ via clustering is critical to improve personalization.

We consider a toy example with linear regression where we have three clients with true models as in \Cref{fig:lrtoy}(a) where true model 0 and 1 are similar to each other but true model 2 is different. The global model trained from vanilla FedAvg does not match well with any true models as shown in \Cref{fig:lrtoy}(a). If we minimize \cref{eqn:main_objective} with respect to $\wb_\ic$ without clustering, i.e., $\sw_{k,i}=1/K,~\forall~k,~i\in[\ec]$, the output local model also diverges from the true model for each client, especially for client 0 and 1, due to the heterogeneity across the true models (see \Cref{fig:lrtoy}(b)-(d)). Finally, if we minimize \cref{eqn:main_objective} with clustering so that for client $k$, higher weight $\sw_{k,i}$ is given to the client $i$ that has a similar true model to client $k$, and smaller weight is given to the other client that has a different true model, the output local model of client $\ic$ gets close to its true model. This is further explored theoretically in \Cref{the2}. \pkt is based on this motivation where we set the weights for co-distillation in $\siglb_\ic(\xb)=\sum_{i=1}^K \sw_{k,i}\sigl(\wb_i,\xb)$ so that each client $k$ sets higher $\sw_{k,i}$, $i\in[\ec]$ for client $i$ that has smaller difference between $\sigl(\wb_i,\xb)$ and $\sigl(\wb_k,\xb)$. Details of the setup for \Cref{fig:lrtoy} are in \Cref{app:exp}.
\subsection{Algorithm.} 
\label{sec:algo}

We minimize \cref{eqn:main_objective} with respect to $\wb_\ic$ for each client $\ic$ on its own device with only communicating the logits instead of the actual model $\wb_\ic$, with the server. With $(t,r)$ denoting the communication round $t$ and local iteration $r$, we define $\siglb_\ic^{(t,0)}=\sum_{i=1}^K \sw_{k,i}^{(t,0)}\siglc_i^{(t,0)}$ for $t\in[0,T-1]$ and $r\in[0,\tau-1]$ where $\siglb_\ic^{(t,0)}$ is fixed for all $r$ and updated only for every $t$. The term $T$ and $\tau$ is the total number of communication rounds and local iterations respectively. Note that the logit information $\siglb_\ic^{(t,0)}$ is computed and sent by the server to the clients for every communication round $t$. Details are in the following paragraphs and \Cref{algo1}.

\paragraph{Client Side Update.} From \cref{eqn:main_objective}, given $\siglb_\ic^{(t,0)}$ from the server, each client's local update rule is:
\begin{align}
&\begin{aligned}
    \wb_\ic^{(t,r+1)}=\wb_\ic^{(t,r)}-\lr&\left[\frac{2\lambda}{|\mathcal{P}_\ic^{(t,r)}|}\sum_{\xb\in\mathcal{P}_\ic^{(t,r)}}\nabla s(\wb_\ic^{(t,r)}, \xb)^T\left( s(\wb_\ic^{(t,r)}, \xb)-\siglb_\ic^{(t,0)}(\xb)\right) \right. \\ 
    &\left.+\frac{1}{|\xi_\ic^{(t,r)}|}\sum_{\xi \in |\xi_\ic^{(t,r)}|}\nabla \lff(\wb_\ic^{(t,r)},\xi)\right] \label{eqn:lcupdate}
    \end{aligned}\\
    &~~\quad\quad\quad\triangleq\wb_\ic^{(t,r)}-\lr\sgrad(\wb_\ic^{(t,r)};\siglb_\ic^{(t,0)})
\end{align}
The term $\wb_\ic^{(t,r)}$ denotes the local model parameters of client $\ic$, $\lr$ is the learning rate, $\xi_\ic^{(t,r)}$ is the mini-batch randomly sampled from client $\ic$'s local dataset $\mathcal{B}_k$, and $\mathcal{P}_\ic^{(t,r)}$ is the mini-batch randomly sampled from the public dataset from client $\ic$. 
We also denote the updated local model of client $k$ after all $\tau$ local iterations for round $t$ as $\wb_\ic^{(t+1,0)}=\wb_\ic^{(t,\tau)}$.

For \pktc, we consider partial client participation where for every communication round $t$, $m$ clients are selected with probability $p_k$ without replacement from $\ic\in[\ec]$. We denote the set of selected clients as $\mathcal{S}^{(t,0)}$ that is fixed for all local iterations $r\in[0,\tau-1]$. If a client $\ic\in[\ec]$ was most recently selected in the previous communication round $t'<t$, and selected again for the current communication round $t$, we assume that $\wb_\ic^{(t,0)}=\wb_\ic^{(t',\tau)}$. In other words, we retrieve the most recently updated local model for the client that is selected for the next communication round and use that model for local training. Each client $\ic\in\mathcal{S}^{(t,0)}$ takes $\tau\geq1$ local updates before sending its logits back to the server where each local update step follows the update in \cref{eqn:lcupdate}. 


\paragraph{Server Side Clustered Knowledge Aggregation.} After the $\tau$ local iterations, each client $\ic\in\mathcal{S}^{(t,0)}$ sends the logits from its updated local model to the server. The logits are denoted as
 $\siglc_\ic^{(t+1,0)}\in\mathbb{R}^{|\mathcal{P}|\times N}$ which is $\sigl(\wb_\ic^{(t+1,0)},\xb)=\sigl(\wb_\ic^{(t,\tau)},\xb)\in\mathbb{R}^{1\times N}$ stacked in to rows for each $\xb\in\mathcal{P}$. The server uses $c$-means clustering (also known conventionally as the $k$-means clustering algorithm \cite{hart1979kmeans}) to cluster the received $m$ different set of logits, $\siglc_\ic^{(t+1,0)},\ic\in\mathcal{S}^{(t)}$ to $c$ clusters, where $c$ is an integer such that $1\leq c\leq m$. The server also gets the next set of selected clients $\mathcal{S}^{(t+1,0)}$ and sends the centroids $\{\cent_i^{(t+1,0)}\}_{i\in[\nc]}$ for each cluster to the clients in $\mathcal{S}^{(t+1,0)}$. Each client $k'\in\mathcal{S}^{(t+1,0)}$ then determines the centroid that is closest to its current model's logit as
 \begin{align}
 \siglb_{\ic'}^{(t+1,0)}=\argmin_{\{\cent_i^{(t+1,0)}\}_{i\in[\nc]}}\|\cent_i^{(t+1,0)}-\siglc_{\ic'}^{(t+1,0)}\|_2^2 \label{eqn:cluster}
 \end{align}
 and uses it for the local update in \cref{eqn:lcupdate}. Since we defined $\siglb_{\ic'}^{(t+1,0)}=\sum_{i=1}^K \sw_{k',i}^{(t+1,0)}\siglc_i^{(t+1,0)}$, \Cref{eqn:cluster} gives a natural selection of $\sw_{k',i}^{(t+1,0)}$ which gives higher weight to the $\siglc_i^{(t+1,0)},~i\in\mathcal{S}^{(t,0)}$ that is closer to client $k'$'s logits, i.e., $\siglc_{k'}^{(t+1,0)}$. 
 \pkt can set $\sw_{k,i}=0$ for client $i$ if its logit is significantly different from the logit of client $k$ or if it was not included in the previous set of selected clients. 

\begin{algorithm} 
\caption{Personalized Federated Learning with Clustered Knowledge Transfer (\pktc)}\label{algo1}
\renewcommand{\algorithmicloop}{\textbf{Global server do:}}
\begin{algorithmic}[1]
\STATE {\bfseries Input:} $m,~\{p_k\}_{\ic\in[\ec]}$, mini-batch size $b$,$b'$ for private, public data each, number of clusters $c$
\STATE {\bfseries Output:} $\{\wb_\ic\}_{\ic\in[\ec]}$
\STATE {\bfseries Initialize:} $\{\siglc_\ic^{(0,0)}\}_{\ic\in\mathcal{S}^{(-1,0)}}$, selected set of $m$ clients $\mathcal{S}^{(-1,0)}$
\STATE {\bfseries For ${t=0,...,T-1}$ communication rounds do}:
\STATE \hspace*{1em} {\bfseries Global server do:}
\STATE \hspace*{2em} Cluster $\{\siglc_\ic^{(t,0)}\}_{\ic\in\mathcal{S}^{(t-1,0)}}$ by $c$-means clustering
\STATE \hspace*{2em} Get centroids $\{\cent_i^{(t,0)}\}_{i\in[\nc]}$ for each cluster 
\STATE \hspace*{2em} Select $m$ clients for $\mathcal{S}^{(t,0)}$ without replacement\\
\hspace*{2em} from $[K]$ by the dataset ratio $\{p_k\}_{\ic\in[\ec]}$ 
\STATE \hspace*{2em} Send centroids $\{\cent_i^{(t,0)}\}_{i\in[\nc]}$ to clients $k\in\mathcal{S}^{(t,0)}$

\STATE \hspace*{1em} {\bfseries Clients $\ic\in\mathcal{S}^{(t,0)}$ in parallel do:}
\STATE \hspace*{2em} {Get $\siglc_\ic^{(t,0)}$ for current local model $\wb_\ic^{(t,0)}$, and find\\ \hspace*{2em} $\siglb_\ic^{(t,0)}=\argmin_{\{\cent_i^{(t,0)}\}_{i\in[\nc]}}\|\cent_i^{(t,0)}-\siglc_\ic^{(t,0)}\|_2^2$.}
\STATE \hspace*{2em} {\bfseries For $r=0,...,\tau-1$ local iterations do:}
\STATE \hspace*{3em} Create mini-batch $\xi_\ic^{(t,r)}$ from sampling $b$ sam-\\
\hspace*{3em} ples uniformly at random from $\mathcal{B}_\ic$, and mini-\\
\hspace*{3em} batch $\mathcal{P}_\ic^{(t,r)}$ from sampling $b'$ samples uniformly\\
\hspace*{3em} at random from $\mathcal{P}$
\STATE \hspace*{3em} Update $\wb_\ic^{(t,r+1)}\leftarrow\wb_\ic^{(t,r)}-\eta\sgrad(\wb_\ic^{(t,r)};\siglb_\ic^{(t,0)})$
\STATE \hspace*{2em} Send $\siglc_\ic^{(t+1,0)}=\siglc_\ic^{(t,\tau)}$ for the updated local model\\
\hspace*{2em} $\wb_\ic^{(t,\tau)}$ back to the server
\end{algorithmic}
\end{algorithm}

We show in the subsequent sections that our proposed \pkt indeed converges and improves the generalization performance of the individual clients' personalized models by clustering. We also empirically show that \pkt achieves high test accuracy as state-of-the-art (SOTA) personalized FL algorithms with drastically smaller communication cost. 

\section{Theoretical Analysis of \pkt} \label{sec:theo}
In this section, we analyze the convergence and generalization properties of \pktc, specifically highlighting the effect of clustered knowledge transfer to generalization. 

\subsection{Convergence Analysis} \label{sec:theoconv}
Here, we present the convergence guarantees of \pkt with regards to the objective function $\gc_\ic(\wb_\ic^{(t,0)};\siglb_\ic^{(t,0)})$ as $t\rightarrow\infty$ with $\tau=1$. 
We use the following assumptions for our analysis:
\begin{assumption}
The composite loss function $f(\wb,\xi)$ is Lipschitz-continuous and Lipschitz-smooth for all $\wb,~\xi$, and therefore $F_1(\wb),~...,~\lf(\wb)$ are all $L_f$-continuous and $L_p$-smooth for all $\wb$. \label{as1}
\end{assumption} 
\begin{assumption}
Each $F_1,~...,~\lf$ is bounded below by a scalar $F_{\ic,\inf}$ over its domain for $\ic\in[\ec]$. \label{as2}
\end{assumption} 
\begin{assumption}
For the mini-batch $\xi_k$ uniformly sampled at random from $\bdat$, the resulting stochastic gradient is unbiased, that is, $\expt\left[\frac{1}{
    |\xi_\ic|}\sum_{\xi \in \xi_\ic}\nabla \lff(\wb_\ic, \xi)\right]=\nabla \lf(\wb_\ic)$. 
\label{as3}
\end{assumption} 
\begin{assumption}
~The stochastic gradient's expected squared norm is uniformly bounded, i.e.,  $\expt\left\|\frac{1}{
    |\xi_\ic|}\sum_{\xi \in \xi_\ic}\nabla \lff(\wb_\ic, \xi)\right\|^2\leq \grad^2$ for $\ic=1,...,\ec$. \label{as4}
\end{assumption} 
\begin{assumption}
$\sigl(\wb,\xb)$ is $L_s-$continuous and $L_g-$smooth for all $\wb$ and $x$. \label{as5}
\end{assumption}
Now we present the convergence guarantees for \pkt in \Cref{the1} below:
\begin{theorem}
With \Cref{as1}-\Cref{as5}, after running \pkt (\Cref{algo1}) for $t=T$ iterations on client $k\in[K]$ with $K$ total clients participating, with the learning rate satisfying $\sum_{t=0}^{\infty}\lr=\infty,~\sum_{t=0}^{\infty}\lr^2<\infty$, we have that the norm of the gradient of $\gc_\ic(\wb_\ic^{(t,0)};\siglb_\ic^{(t,0)})$ with respect to $\wb_\ic^{(t,0)}$ given $\siglb_\ic^{(t,0)}$ goes to zero with probability 1 as $T\rightarrow\infty$, i.e., for every client $k$,
\begin{align}
\lim_{t\rightarrow\infty} \|\nabla_{\wb_\ic^{(t,0)}}\gc_\ic(\wb_\ic^{(t,0)};\siglb_\ic^{(t,0)})\|=0
\end{align} \label{the1}
\end{theorem}
The proof for \Cref{the1} is presented in the \Cref{app:the1proof}. \Cref{the1} shows that our proposed algorithm \pkt indeed converges to a first-order stationary point with respect to $\wb_\ic$ given $\siglb_\ic$ where the norm of the gradient of our main objective function $\gc_\ic(\wb_\ic;\siglb_\ic)$ with respect to $\wb_\ic$ is 0. 

\subsection{Generalization Performance} \label{sec:theoclust}
Now we show the theoretical grounds for clustered knowledge distillation in regards to the generalization performance for personalized FL in the problem of linear regression. We also present a generalization bound for ensemble models in the context of personalization in \Cref{app:genbound}. For $K$ clients in total, we consider a Bayesian framework as in \cite{li2021ditto} where we have $\theta$ uniformly distributed on $\mathbb{R}^d$, and each device has its data distributed with parameters $\wb_\ic=\theta+\zeta_\ic$ where $\zeta_\ic\sim\mathcal{N}(0,\upsilon_k^2\ib_d)$ and $\mathbf{I}_d$ is the $d\times d$ identity matrix and $\upsilon_\ic$ is unique to the client's task. Suppose we have $\yb_\ic=\xxb_\ic\wb_\ic+\zb,~\ic\in[K]$ where $\yb_\ic\in\mathbb{R}^n,~\xxb_\ic\in\mathbb{R}^{n\times d}$, and $\zb\in\mathbb{R}^n$ such that $\zb\sim\mathcal{N}(0,\sigma^2\ib_d)$. 

Let us consider a linear regression problem for each device $k$ such that we have the empirical loss function as $F_\ic(\wb_\ic)=\|\xxb_\ic\wb_\ic-\yb_\ic\|_2^2$. We have that $\wbt_\ic=(\xxb_\ic^T\xxb_\ic)^{-1}\xxb_\ic^T\yb_\ic$ is a noisy observation of $\wb_\ic$ with additive covariance $\sigma^2(\xxb_\ic^T\xxb_\ic)^{-1}$ since $\wbt_\ic\sim\mathcal{N}((\xxb_\ic^T\xxb_\ic)^{-1}\xxb_\ic^T\yb_\ic,\sigma^2(\xxb_\ic^T\xxb_\ic)^{-1})$. Then using Lemma 2 from \cite{li2021ditto}, with the following definitions:
\begin{align}
    \Sigma_\ic\coloneqq\sigma^2(\xxb_\ic^T\xxb_\ic)^{-1}+\upsilon_\ic^2\ib_d\\
    \overline{\Sigma}_{\setminus\ic}=\left(\sum_{i\in[\ec],i\neq \ic}\Sigma_i^{-1}\right)^{-1}\\
    \overline{\theta}_{\setminus \ic}\coloneqq\overline{\Sigma}_{\setminus\ic}\sum_{i\in[\ec],i\neq \ic}\Sigma_i^{-1}\wbt_i
\end{align}
given $\{\xxb_i,\yb_i\}_{i\in[K],i\neq k}$ we have that 
\begin{align}
    \theta=\overline{\theta}_{\setminus \ic}+\gamma
\end{align}
where $\gamma\sim\mathcal{N}(0,\overline{\Sigma}_{\setminus\ic})$. Further, if we let 
\begin{align}
    &\widetilde{\Sigma}_\ic\coloneqq\overline{\Sigma}_{\setminus\ic}+\upsilon_\ic^2\ib_d\\
    &\overline{\Sigma}_\ic\coloneqq\left((\widetilde{\Sigma}_\ic)^{-1}+(\sigma^2(\xxb_\ic^T\xxb_\ic)^{-1})^{-1}\right)^{-1}
\end{align}
given $\{\xxb_i,\yb_i\}_{i\in[K]}$, again with Lemma 2 from \cite{li2021ditto} we have
\begin{align}
    \wb_\ic= \overline{\Sigma}_\ic (\sigma^2(\xxb_\ic^T\xxb_\ic)^{-1})^{-1}\wbt_\ic+\overline{\Sigma}_\ic( \widetilde{\Sigma}_\ic)^{-1}\overline{\theta}_{\setminus \ic}+\vartheta_\ic
\label{eq22-0}
\end{align}
where $\vartheta_\ic\sim\mathcal{N}(0,\overline{\Sigma}_\ic)$. The term for $\wb_\ic$ in \cref{eq22-0} uses the fact that $\wbt_\ic$ is a noisy observation of $\wb_\ic$ with additive noise of zero mean and covariance $\sigma^2(\xxb_\ic^T\xxb_\ic)^{-1}$, and $\overline{\theta}_{\setminus \ic}$ is a noisy observation of $\theta$ with covariance $\overline{\Sigma}_{\setminus\ic}$. Given all the training samples from $K$ devices, $\wb_\ic$ in \cref{eq22-0} is Bayes optimal. 

With \pktc, following \cref{eqn:main_objective}, we solve the following objective:
\begin{align}
    \min_{\wb_\ic}\|\xxb_\ic\wb_\ic-\yb_\ic\|_2^2+\lambda_\ic\|\siglb_\ic-\sigl(\wb_\ic)\|_2^2 \label{eq:lr}
\end{align}
where $\lambda_\ic$ is the regularization term as in \cref{eqn:main_objective} and $\siglb_\ic$ and $\sigl(\wb_\ic)$ each is comparative to the $\siglb_\ic(\xb)$ and $\sigl(\wb_\ic,\xb)$ in \cref{eqn:main_objective} for a single public data point $\xb$. Note that in the setting of linear regression we can set $\sigl(\wb_\ic)=\pb\wb_\ic$ where $\pb\in\mathbb{R}^{1\times d}$ is the public data (without loss of generality, we assume single data point for the public data for simplicity). Accordingly, we set $\siglb_\ic=\sum_{i=1}^K\sw_{\ic,i}\sigl(\wbt_i)$ for an arbitrary set of weights $\sw_{\ic,i},i\in[K]$ for client $\ic$. Then we have that the local empirical risk minimizer for \cref{eq:lr} is
\begin{align}
\begin{aligned}
    \wbl_\ic=(\xxb_\ic^T\xxb_\ic+\lambda_\ic\pb^T\pb)^{-1}(\xxb_\ic^T\xxb_\ic\wbt_\ic+\lambda_\ic\pb^T\pb\sum_{i=1}^K\sw_{\ic,i}\wbt_i) \label{eq:lr1}
    \end{aligned} 
\end{align}
Finally, we present the optimal $\lambda_\ic^*$ and $\sw_{\ic,i}^*$ for any device $k\in[K]$ given the above linear regression problem with \pkt in \Cref{the2}.
\begin{theorem}
Assuming $\xxb_\ic^T\xxb_\ic=\beta\ib_d$ and $\pb^T\pb=\nu\ib_d$ for some constant $\beta,\nu$, the $\lambda_\ic^*$ and $\sw_{\ic,i}^*,i\in[K]$ that minimizes the test performance on device $\ic,~\ic\in[K]$ i.e.,
\begin{align}
    \lambda_\ic^*,\sw_{\ic,i}^*,i\in[K]=\argmin_{\lambda_\ic,\sw_{\ic,i},i\in[K]}\expt[F_\ic(\wbl_\ic)|\wbt_\ic,\overline{\theta}_{\setminus \ic}]
\end{align}
we have that 
\begin{align}
    \lambda_\ic^*=\sigma^2/\upsilon_\ic^2\nu,~\sw_{\ic,i}^*=\frac{B_k}{\sigma^2+\beta\upsilon_i^2} \label{eqn:lropt}
\end{align}
with $A_k=\left(\sum_{i\in[K],i\neq\ic}\frac{1}{\sigma^2+\beta\upsilon_i^2}\right)^{-1},~B_k=\frac{A_k(\sigma^2+\beta\upsilon_\ic^2)}{\sigma^2+A_k\beta\upsilon_\ic^2}$. \label{the2}
\end{theorem}
\Cref{the2} shows that given the objective function in \cref{eq:lr} and the corresponding minimizer \cref{eq:lr1}, in a data-heterogeneous scenario where $\upsilon_\ic,~\ic\in[\ec]$ is unique to each client $k$, we have that the optimal weights $\sw_{\ic,i}^*,~i\in[K]$ for client $k$ is in fact inversely proportional to $\upsilon_i$. Intuitively, this means that since larger $\upsilon_i$ leads to a larger divergence from the original $\theta$ in $\wb_i=\theta+\zeta_i$, giving lower weight $\sw_{\ic,i}$ to client $i$ improves generalization of the personalzied model. This gives new insight into co-distillation for personalization in FL since previous work \cite{bist2020distdis, li2019fedmd} only consider scenarios where the weight $\sw_{i,\ic}=1/K,~\forall~i,\ic\in[K]$ in a non-personalized FL setting. The results also present strong motivation for clustered knowledge transfer for personalized FL. The proof for \Cref{the2} is presented in \Cref{app:the2proof}. Further discussions on the implications of \Cref{the2} is presented in \Cref{app:the2dis}.

\section{Experiments} \label{sec:expall}
For all experiments we randomly sample a fraction $(C)$ of clients from $K=100$ clients per communication round for local training. For the sake of simplicity and fair comparison across different benchmarks, we use do not apply any momentum acceleration or weight decay to local training. Further details of the experimental setup are in \Cref{app:exp}.

\subsection{Experimental Setup}
\begin{wrapfigure}{r}{0.5\textwidth}
\vspace{-1.5em}
\centering
\begin{subfigure}{0.16\textwidth}
\centering
\includegraphics[width=1\textwidth]{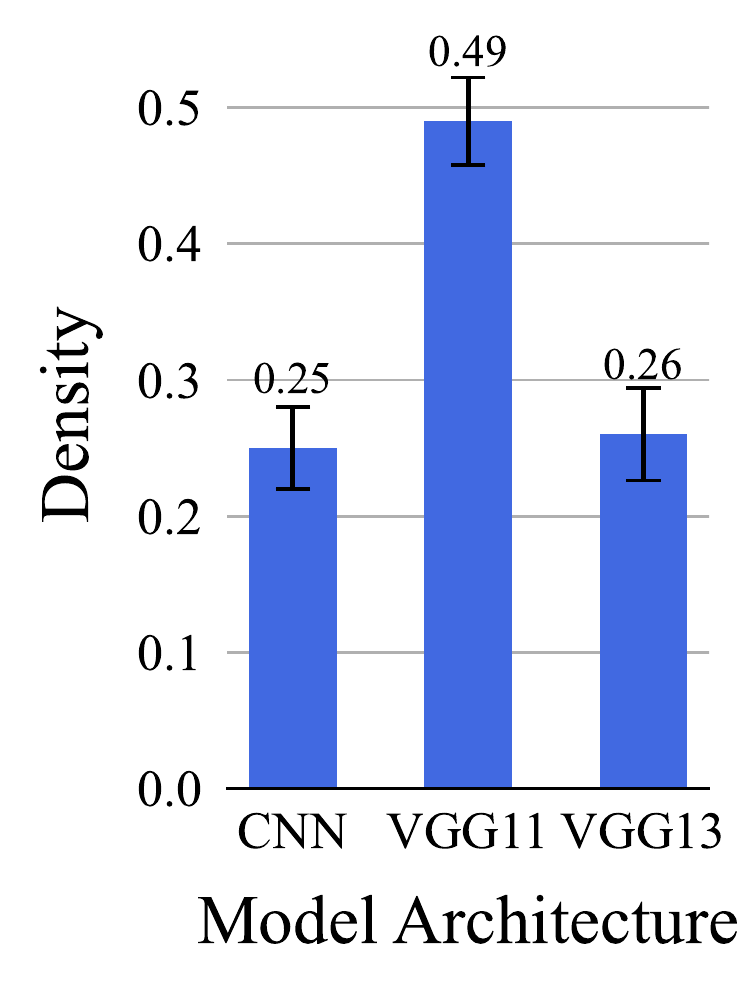}  \vspace{-1.8em}
\caption{}
\end{subfigure}
\begin{subfigure}{0.28\textwidth}
\centering
\includegraphics[width=1\textwidth]{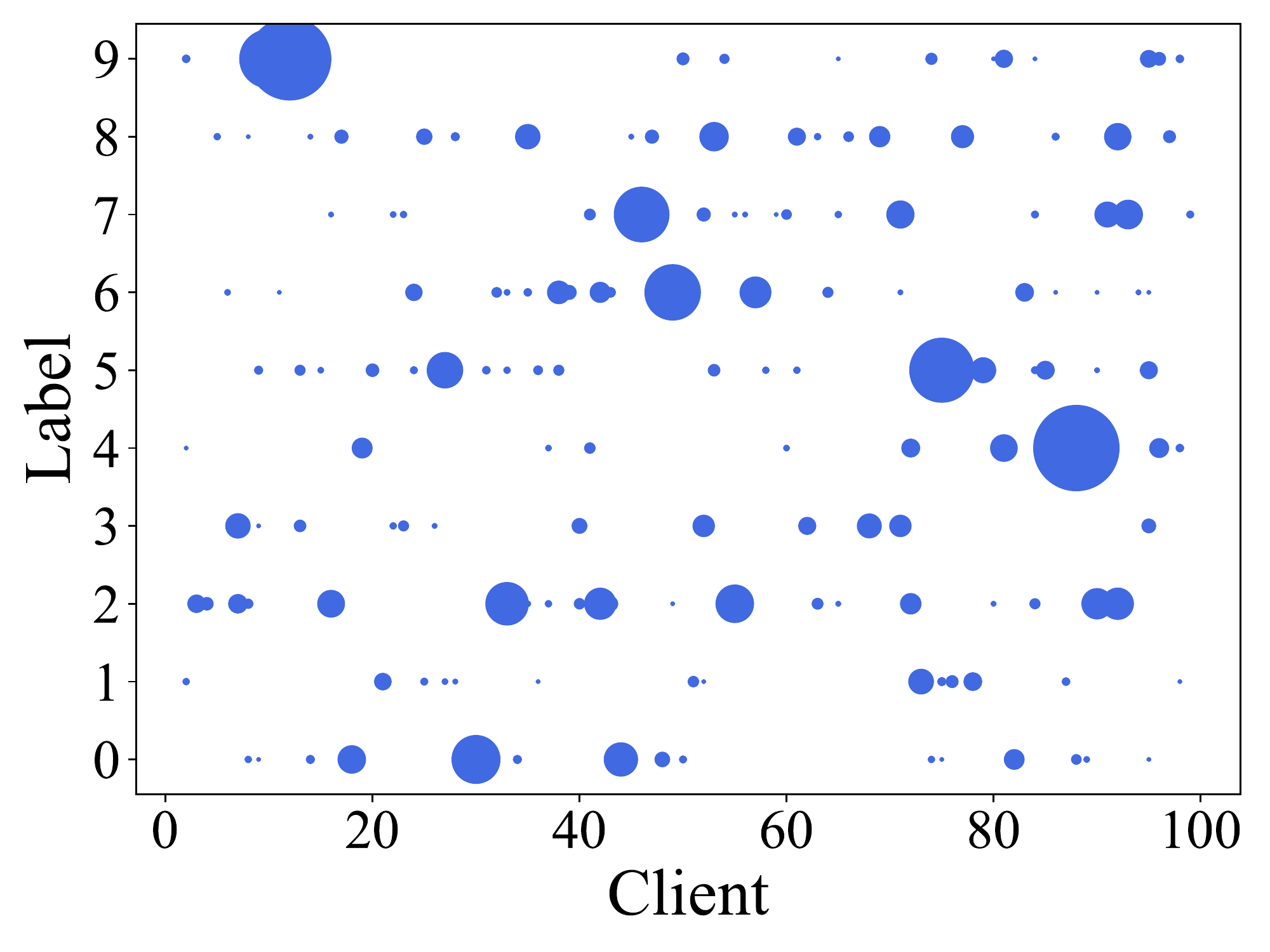} \vspace{-1.67em}
\caption{}
\end{subfigure}
\vspace{-0.5em}
\caption{(a) Proportion of the models architectures deployed across clients for the model heterogeneity scenario; (b) data-distribution with $\alpha=0.01$ for all clients where larger circle indicates larger dataset size for each label 0-9 of CIFAR10.}
  \label{fig:clidetail} 
\vspace{-3.3em}
\end{wrapfigure}

\paragraph{Datasets and models.} We evaluate \pkt with CIFAR10~\cite{krizhevsky2009cifar} as the training/test dataset and CIFAR100~\cite{kri2009cifar100} as the public dataset for image classification in mainly two different scenarios: model homogeneity and heterogeneity. For model homogeneity, VGG11~\cite{simonyan2015very} is deployed for all clients. For model heterogeneity, we sample one of VGG13/VGG11/CNN model architecture for each client with the probability of a larger model getting assigned to a client is proportional to the client's dataset size (see~\Cref{fig:clidetail}(a)). We partition data heterogeneously amongst clients using the Dirichlet distribution $\text{Dir}_{K}(\alpha)$ \citep{hsu2019noniid}, smaller $\alpha$ leads to higher data size imbalance and degree of label skew across clients. We set $\alpha=0.01$ to emulate realistic FL scenarios with large data-heterogeneity (see~\Cref{fig:clidetail}(b)). 

\paragraph{Baselines.} We compare \pkt with SOTA FL algorithms designed to efficiently train either (i) a single global model at the server (e.g. FedAvg, FedProx, Scaffold, FedDF) or (ii) personalized model(s) either at the server side as a global model (GM) or client side (e.g., Per-FedAvg, FedFomo, Ditto, HypCluster) as a local model (LM). Note that we do not assume a \textit{labeled} public dataset, and instead relax the condition to a \textit{small}\footnote{We only use 2000 unlabeled data-samples per experiment that are sampled uniformly at random from CIFAR100.} and \textit{unlabeled} public dataset and therefore exclude comparison to methods which require training directly on a \textit{labeled} public dataset (e.g., FedMD).


\begin{table*}[!h] \centering 

\small
\caption{Average test accuracy across the entire clients and total communication cost (number of parameters communicated per round) for the model-homogeneous scenario with total number of clients $\ec=100$. The standard deviation for the test accuracy across random seeds is shown in the parenthesis.}
\begin{tabular}{@{}llcccc@{}}\toprule \label{fig:table1}
& & \multicolumn{2}{c}{$C=0.10$} & \multicolumn{2}{c}{$C=0.15$}\\
\cmidrule(lr){3-5} \cmidrule(lr){5-6}
Method & Algorithm & Test Acc. & Com.-Cost & Test Acc. & Com.-Cost \vspace*{-0.2em} \\ \midrule
Local Training & - & $64.02~\textcolor{darkgray}{(\pm0.48)}$ & - & $64.02~\textcolor{darkgray}{(\pm0.48)}$ & - \vspace*{-0.1em} \\ \hline
\multirow{4}{*}{Non-
Personalized} & FedAvg & $15.53~\textcolor{darkgray}{(\pm1.42)}$ & \tikzmark{a}  & $20.00~\textcolor{darkgray}{(\pm2.66)}$ & \tikzmark{c}\vspace*{-0.1em}\\ \cline{2-3}\cline{5-5} \rule{0pt}{2.1ex}
& FedProx & $13.64~\textcolor{darkgray}{(\pm1.23)}$ &  & $16.71~\textcolor{darkgray}{(\pm1.79)}$ & \vspace*{-0.1em} \\ \cline{2-3}\cline{5-5} \rule{0pt}{2.1ex}
& Scaffold & $12.92~\textcolor{darkgray}{(\pm1.46)}$ &  & $16.41~\textcolor{darkgray}{(\pm1.14)}$ & \vspace*{-0.1em} \\ \cline{2-3}\cline{5-5} \rule{0pt}{2.1ex}
& FedDF & $15.18~\textcolor{darkgray}{(\pm1.18)}$ & & $17.09~\textcolor{darkgray}{(\pm1.54)}$ & \vspace*{-0.1em} \\ \cline{1-3}\cline{5-5} \rule{0pt}{2.1ex}
\multirow{13}{*}{Personalized} & Per-FedAvg (GM) & $14.47~\textcolor{darkgray}{(\pm0.59)}$  &  & $14.61~\textcolor{darkgray}{(\pm0.83)}$ & \vspace*{-0.1em} \\ \cline{2-3}\cline{5-5} \rule{0pt}{2.1ex}
& Per-FedAvg (LM) & $51.74~\textcolor{darkgray}{(\pm1.52)}$ &  & $50.25~\textcolor{darkgray}{(\pm1.61)}$ & \vspace*{-0.1em} \\ \cline{2-3}\cline{5-5} \rule{0pt}{2.1ex}
& Ditto (LM) & $67.21~\textcolor{darkgray}{(\pm1.86)}$ & & $68.88~\textcolor{darkgray}{(\pm1.95)}$ & \vspace*{-0.1em} \\ \cline{2-3}\cline{5-5} \rule{0pt}{2.1ex}
& Ditto (GM) & $21.63~\textcolor{darkgray}{(\pm2.13)}$ & \tikzmark{b} & $19.16~\textcolor{darkgray}{(\pm2.34)}$ & \tikzmark{d}  \vspace*{-0.1em} \\ \cline{2-6} \rule{0pt}{2.1ex}
& FedFomo & \textbf{74.62}~\textcolor{darkgray}{($\pm0.42$)} & $\bm{\mathrm{3900\times10^7}}$ & \textbf{77.56}~\textcolor{darkgray}{($\pm0.75$)} & $\bm{\mathrm{5850}\times10^7}$ \vspace*{-0.1em} \\  \cline{2-6} \rule{0pt}{2.1ex}
& HypCluster ($c=2$) & $34.11~\textcolor{darkgray}{(\pm2.63)}$ & $2340\times10^7$ & $28.70~\textcolor{darkgray}{(\pm3.03)}$ & $3320\times10^7$ \vspace*{-0.1em} \\ \cline{2-6} \rule{0pt}{2.1ex}
& HypCluster ($c=3$) & $39.17~\textcolor{darkgray}{(\pm2.64)}$ & $2540\times10^7$ & $41.99~\textcolor{darkgray}{(\pm2.48)}$ & $3510\times10^7$ \vspace*{-0.1em} \\ \cline{2-6} \rule{0pt}{2.1ex}
& HypCluster ($c=5$) & $52.51~\textcolor{darkgray}{(\pm1.37)}$ & $2930\times10^7$ & $51.92~\textcolor{darkgray}{(\pm1.64)}$ & $3900\times10^7$ \vspace*{-0.1em} \\ \cline{2-6} \rule{0pt}{2.1ex}
& HypCluster ($c=6$) & $65.77~\textcolor{darkgray}{(\pm2.76)}$ & $3120\times10^7$ & $63.28~\textcolor{darkgray}{(\pm1.16)}$ & $4100\times10^7$ \vspace*{-0.1em} \\ \cline{2-6} \rule{0pt}{2.1ex}
& \pkt ($c=1$) & $70.70~\textcolor{darkgray}{(\pm0.46)}$ & $4.4\times10^7$ & $67.66~\textcolor{darkgray}{(\pm0.30)}$ & $6.4\times10^7$ \vspace*{-0.1em} \\ \cline{2-6} \rule{0pt}{2.1ex}
& \pkt ($c=2$) & $73.33~\textcolor{darkgray}{(\pm0.26)}$ & $4.8\times10^7$ & $70.86~\textcolor{darkgray}{(\pm0.73)}$ & $6.8\times10^7$ \vspace*{-0.1em} \\  \cline{2-6} \rule{0pt}{2.1ex}
& \pkt ($c=3$) & \textbf{74.31} \textcolor{darkgray}{($\pm0.40$)} & $\bm{\mathrm{5.2\times10^7}}$ & \textbf{76.74}~\textcolor{darkgray}{($\pm0.71$)} & $\bm{\mathrm{7.2\times10^7}}$ \vspace*{-0.1em} \\  \cline{2-6} \rule{0pt}{2.1ex}
& \pkt ($c=4$) & $72.67~\textcolor{darkgray}{(\pm0.31)}$ & $5.6\times10^7$ & $73.52~\textcolor{darkgray}{(\pm1.15)}$ & $7.6\times10^7$ \vspace*{-0.1em} \\
\bottomrule \vspace{-1em} \label{tab:testcomp}
\end{tabular}

\tikz[remember picture,overlay]
{\draw[double] ([yshift=1ex]a) -- (b) node[midway,fill=white] {$2150\times10^7$};}

\tikz[remember picture,overlay]
{\draw[double] ([yshift=1ex]c) -- (d) node[midway,fill=white] {$3120\times10^7$};}

\end{table*}


\begin{table*}[!h] \centering 

\small
\caption{Average test accuracy across all clients and total communication cost (number of parameters communicated per round) for the model-heterogeneous scenario with total number of clients $\ec=100$. The standard deviation for the test accuracy across random seeds is shown in the parenthesis.}

\begin{tabular}{@{}llcccc@{}}\toprule \label{fig:table2}
& & \multicolumn{2}{c}{$C=0.10$} & \multicolumn{2}{c}{$C=0.15$}\\
\cmidrule(lr){3-5} \cmidrule(lr){5-6}
Method & Algorithm & Test Acc. & Com.-Cost & Test Acc. & Com.-Cost \\ \midrule
Local Training & - & $59.29~\textcolor{darkgray}{(\pm1.05)}$ & - & $59.29~\textcolor{darkgray}{(\pm1.05)}$ & - \vspace*{-0.1em} \\ \hline \rule{0pt}{2.1ex}
{Non-Personalized} & FedDF & $15.25~\textcolor{darkgray}{(\pm1.21)}$ & $1780\times10^7$ & $14.16~\textcolor{darkgray}{(\pm1.11)}$ & $2620\times10^7$ \vspace*{-0.1em} \\  \hline \rule{0pt}{2.1ex}
\multirow{4}{*}{Personalized} 
& \pkt ($c=1$) & $70.94~\textcolor{darkgray}{(\pm0.46)}$ & $4.4\times10^7$ & $70.77~\textcolor{darkgray}{(\pm0.14)}$ & $6.4\times10^7$ \vspace*{-0.1em} \\ \cline{2-6} \rule{0pt}{2.1ex}
& \pkt ($c=2$) & \textbf{72.25}~\textcolor{darkgray}{($\pm0.17$)} & $\bm{\mathrm{4.8\times10^7}}$ & \textbf{76.14}~\textcolor{darkgray}{($\pm0.73$)} & $\bm{\mathrm{6.8\times10^7}}$ \vspace*{-0.1em} \\  \cline{2-6} \rule{0pt}{2.1ex}
& \pkt ($c=3$) & $71.84 \textcolor{darkgray}{(\pm0.40)}$ & $5.2\times10^7$ & 76.02~\textcolor{darkgray}{($\pm0.66$)} & $7.2\times10^7$ \vspace*{-0.1em} \\  \cline{2-6} \rule{0pt}{2.1ex}
& \pkt ($c=4$) & $70.01~\textcolor{darkgray}{(\pm0.31)}$ & $5.6\times10^7$ & $73.14~\textcolor{darkgray}{(\pm0.61)}$ & $7.6\times10^7$ \vspace*{-0.1em} \\
\bottomrule \vspace{-1em} \label{tab:modelhet}
\end{tabular}
\end{table*}

\subsection{Experimental Results} \label{sec:expresults}
We demonstrate the efficacy of \pkt in terms of the average test accuracy across all clients with the communication cost defined as the total number of parameters communicated across server/clients during the training process including uplink and downlink. 

\paragraph{Test Accuracy and Communication Cost.} In \Cref{fig:table1}, we show the performance of \pkt along with the performance of other SOTA FL algorithms in regards to the achieved highest test accuracy and communicated number of parameters between server and client with different fractions of selected clients $C$. For $C=0.1$, we show that \pkt achieves high test accuracy of $74.31\%$ with $c=3$, with small communication cost compared to other algorithms (saving at maximum $\times750$). FedFomo achieves a slightly higher test accuracy performance with $74.62\%$, but the communication cost spent, $3900\times 10^7$ parameters, is significantly larger compared to \pkt which is $5.2\times 10^7$ parameters. Moreover note that algorithms that train a single global model in the non-personalized FL setting performs worse than personalized algorithms showing that the traditional FL framework does not perform well to individual clients in the setting of high data heterogeneity. For $C=0.15$, we also show that \pkt is able to achieve a comparable high test accuracy of $76.74\%$ with only a small communication cost of $7.2\times 10^7$ parameters (saving at maximum $\times812.5$) where FedFomo achieves a slightly higher accuracy of $77.56\%$ with large communication cost of $5850\times 10^7$.

\paragraph{Model Heterogeneity.}
We demonstrate the performance of \pkt where clients have different models dependent on their dataset size (see \Cref{fig:clidetail}(b)) in \Cref{fig:table2}. Note that this is a realistic setting of FL where clients can have smaller or larger models dependent on their dataset size or system capabilities. Only FedDF is capable for model heterogeneity amongst the SOTA FL algorithms which we included for comparison. With model heterogeneity, \pkt achieves high test accuracy $72.25\%$ for $C=0.1$ and $76.14\%$ for $C=0.15$ with even smaller communication cost of $4.8\times10^7$ and $6.8\times10^7$ respectively. For $C=0.15$, the test accuracy is close to that of \pkt for model homogeneity, showing that allowing model heterogeneity increases feasibility while not hurting the local performance of clients.

\paragraph{Effect of Clustering.}
\pkt performs clustering at the server side to cluster the logit information received from the clients. We evaluate how the number of clusters effects the test accuracy and communication cost. For both $C=0.1$ and $C=0.15$ in \Cref{fig:table1}, $c=3$ achieves the best test accuracy performance. Increasing $c$ from that point actually deteriorates the performance with higher communication cost. This shows that while clustering can help to a certain extent, too much clustering can hurt generalization since we are decreasing the number of clients in each cluster and the diversity of information within each cluster. Similar behavior is observed in \Cref{fig:table1} where the best test accuracy is achieved in $c=2$, and then the accuracy decreases for higher $c$.

\section{Concluding Remarks}
The inherently high data and system heterogeneity across resource-constrained clients in FL should be considered for devising realistic personalized FL schemes. However, previous work in personalized FL restricted clients to have homogeneous models across clients with direct communication of the model parameters which can incur heavy communication cost. We propose \pkt that caters to the data and system heterogeneity across clients by using clustered knowledge transfer, allowing heterogeneous model deployment without direct communication of the models. We show that \pkt achieves competitive performance compared to other SOTA personalized FL schemes at a much smaller communication cost. Interesting future directions of this work include understanding the privacy implications that arise due to the clustering of clients and communicating their logits instead of the actual model parameters. 

\newpage
\bibliography{dist_sgd}
\bibliographystyle{unsrt}

\newpage
\appendix

\section{Proof for \Cref{the1}} \label{app:the1proof}
In this section we present the proof for \Cref{the1}. We follow the techniques presented by \cite{bist2020distdis} for the proof. For notational simplicity, we notate all super subscript $(t,0)$ as $(t)$ throughout the proof, dropping the local iteration index. We define the following $\sigma$-algebra on the set that contains the history of the model updates for all clients with 
$\wb^{\ti}=[\wb_1^{\ti}...\wb_\ec^{\ti}]$ and $\siglb^{\ti}=[\siglb_1^{\ti}...\siglb_\ec^{\ti}]$ as $\mathcal{H}_t=\sigma(\{\wb^{(i)},\siglb^{(i)}\}~|~i\leq t)$. Recall $\sgrad(\wb_\ic^{\ti};\siglb_\ic^{\ti})\triangleq\frac{1}{
    |\midat|}\sum_{\xi \in \midat}\nabla \lff(\wb_\ic^{\ti}, \xi) +\frac{2\lambda}{|\mathcal{P}_\ic^{\ti}|}\sum_{\xb\in\mathcal{P}_\ic^{\ti}}\nabla s(\wb_\ic^{\ti}, \xb)^T\left( s(\wb_\ic^{\ti}, \xb)-\siglb_\ic^{(t)}(\xb)\right)$. 

\subsection{Additional Lemmas}
We first present useful Lemmas and their proofs which we use for the intermediate steps in the main proof for \Cref{the1}.

\begin{lemma}
The gradient of the second term in $\gc_\ic(\wb_\ic^{\ti};\siglb_\ic^{\ti})$ with respect to $\wb_\ic^{(t)}$ is Lipschitz continuous, and therefore with \cref{as1}, $\gc_\ic(\wb_\ic^{\ti};\siglb_\ic^{\ti})$ is also a Lipschitz-smooth function with factor $L_p$. \label{lem1}
\begin{proof}
With dropping the iteration index $t$ for the upper script for simplicity, let's define the second term in $\gc_\ic(\wb_\ic;\siglb_\ic)$ as $\q(\wb_\ic;\siglb_\ic)\triangleq\frac{\lambda}{|\mathcal{P}|}\sum_{\xb\in\mathcal{P}}\|\siglb(\xb)-\sigl(\wb_\ic,\xb)\|_2^2$. Then we have that 
\begin{align}
\nabla_{\wb_\ic} \q(\wb_\ic;\siglb_\ic)=\frac{2\lambda}{|\mathcal{P}|}\sum_{\xb\in\mathcal{P}}\nabla s(\wb_\ic, \xb)^T\left( s(\wb_\ic, \xb)-\siglb_\ic(\xb)\right)
\end{align}
For an arbitrary $\mathbf{e}_\ic$ in the domain of $\q(\cdot;\siglb_\ic)$ for each $\xb\in\mathcal{P}$, we have that
\begin{flalign}
&\begin{aligned}
&\|\nabla s(\wb_\ic, \xb)^T\left( s(\wb_\ic, \xb)-\siglb_\ic(\xb)\right)-\nabla s(\eb_\ic, \xb)^T( s(\eb_\ic, \xb)-\siglb_\ic(\xb))\|^2 \\ 
&\leq 3\|\nabla s(\wb_\ic, \xb)^T\left( s(\wb_\ic, \xb)-\siglb_\ic(\xb)\right)-\nabla s(\wb_\ic, \xb)^T\left( s(\eb_\ic, \xb)-\siglb_\ic(\xb)\right)\|^2\\
&\quad+3\|\nabla s(\wb_\ic, \xb)^T( s(\eb_\ic, \xb)-\siglb_\ic(\xb))-\nabla s(\eb_\ic, \xb)^T\left( s(\eb_\ic, \xb)-\siglb_\ic(\xb)\right)\|^2 \label{eq2-0}
\end{aligned}
\\
&\leq 3\|\nabla s(\wb_\ic, x)\|^2\|s(\wb_\ic, x)- s(\eb_\ic, x)\|^2+3\|\nabla s(\wb_\ic, x)-\nabla s(\eb_\ic, x)\|^2\|s(\eb_\ic, x)-\siglb_\ic(x)\|^2 \label{eq2-1}
\\
&\leq3L_s^4\|\wb_\ic-\eb_\ic\|^2+6L_g^2\|\wb_\ic-\eb_\ic\|^2 \label{eq2-2}
\\
&=(3L_s^4+6L_g^2)\|\wb_\ic-\eb_\ic\|^2 
\end{flalign}
where \cref{eq2-0} uses Jensen's inequality for the \(\ell_2\)-norm for three terms, \cref{eq2-1} uses the submultiplicativity of the norm, and the LHS of \cref{eq2-2} uses \cref{as5}. Therefore we can conclude that $\|\nabla s(\wb_\ic, \xb)^T( s(\wb_\ic, \xb)-\siglb_\ic(\xb))-\nabla s(\eb_\ic, \xb)^T( s(\eb_\ic, \xb)-\siglb_\ic(\xb))\|^2$ for any $\xb$ is Lipschitz-continuous, and hence $\nabla_{\wb_\ic} \q(\wb_\ic;\siglb_\ic)$ is also Lipschitz-continuous.
\end{proof}
\end{lemma}
\begin{lemma} 
We have that $\expt[\sgrad(\wb_\ic^{\ti};\siglb_\ic^{\ti})|\mathcal{H}_t]=\nabla_{\wb_\ic^{\ti}}\gc_\ic(\wb_\ic^{\ti};\siglb_\ic^{\ti})$ and $\expt[\|\sgrad(\wb_\ic^{\ti};\siglb_\ic^{\ti})\|_2^2]\leq 2G^2+16\lambda^2L_s^2$ and $\|\nabla_{\wb_\ic^{\ti}}\gc_\ic(\wb_\ic^{\ti};\siglb_\ic^{\ti})\|$ and $\|\sgrad(\wb_\ic^{\ti};\siglb_\ic^{\ti})\|$ is each bounded by constant $M_1\geq0$ and $M_2\geq0$. \label{lem2}
\begin{proof}
By definition of the gradient we have
\begin{align}
&\begin{aligned}
&\expt[\sgrad(\wb_\ic^{\ti};\siglb_\ic^{\ti})|\mathcal{H}_t]\\
&=\expt[\frac{1}{
    |\midat|}\sum_{\xi \in \midat}\nabla \lff(\wb_\ic^{\ti}, \xi) +\frac{2\lambda}{|\mathcal{P}_\ic^{\ti}|}\sum_{\xb\in\mathcal{P}_\ic^{\ti}}\nabla s(\wb_\ic^{\ti}, \xb)^T( s(\wb_\ic^{\ti}, \xb)-\siglb_\ic^{(t)}(\xb))|\mathcal{H}_t]
\end{aligned}
\\
&\begin{aligned}
=\nabla \lf(\wb_\ic^{\ti})+ \frac{2\lambda}{|\mathcal{P}|}\sum_{\xb\in\mathcal{P}}\nabla s(\wb_\ic^{\ti}, \xb)^T(s(\wb_\ic^{\ti}, \xb)
-\siglb_\ic^{(t)}(\xb))
\end{aligned}\\
&=\nabla_{\wb_\ic^{\ti}}\gc_\ic(\wb_\ic^{\ti};\siglb_\ic^{\ti})
\end{align}
finishing the proof for the first part of \cref{lem2}. Next, we prove the second part of \cref{lem2} showing that
\begin{flalign}
&\begin{aligned}
&\expt[\|\sgrad(\wb_\ic^{\ti};\siglb_\ic^{\ti})\|^2]\\&=\expt[\|\frac{1}{|\midat|}\sum_{\xi \in \midat}\nabla \lff(\wb_\ic^{\ti}, \xi) + \frac{2\lambda}{|\mathcal{P}_\ic^{\ti}|}\sum_{\xb\in\mathcal{P}_\ic^{\ti}}\nabla s(\wb_\ic^{\ti}, \xb)^T( s(\wb_\ic^{\ti}, \xb)-\siglb_\ic^{(t)}(\xb))\|^2]
\end{aligned} \\
&\leq 2\expt[\|\frac{1}{
    |\midat|}\sum_{\xi \in \midat}\nabla \lff(\wb_\ic^{\ti}, \xi)\|^2]+2\expt[\|\frac{2\lambda}{|\mathcal{P}_\ic^{\ti}|}\sum_{\xb\in\mathcal{P}_\ic^{\ti}}\nabla s(\wb_\ic^{\ti}, \xb)^T( s(\wb_\ic^{\ti}, \xb)-\siglb_\ic^{(t)}(\xb))\|^2] \label{eq2-3}
\\
& \leq \expt[\frac{8\lambda^2}{|\mathcal{P}_\ic^{\ti}|}\sum_{\xb\in\mathcal{P}_\ic^{\ti}}\|\nabla s(\wb_\ic^{\ti}, \xb)^T(s(\wb_\ic^{\ti}, \xb)-\siglb_\ic^{(t)}(\xb))\|^2]+2G^2 \label{eq2-4}
\\
&\leq \frac{8\lambda^2}{|\mathcal{P}|}\sum_{\xb\in\mathcal{P}}\expt[\|\nabla s(\wb_\ic^{\ti}, \xb)\|^2\| s(\wb_\ic^{\ti}, \xb)-\siglb_\ic^{(t)}(\xb)\|^2]+2G^2\label{eq2-5}
\\
&\leq 2G^2+16\lambda^2L_s^2 \label{eq2-6}
\end{flalign}
where \cref{eq2-3} is due to the Cauchy–Schwarz inequality and AM-GM inequality, \cref{eq2-4} is due to \cref{as4} and Jensen's inequality, \cref{eq2-5} is due to the submultiplicativity of the norm, and \cref{eq2-6} is due to \cref{as5} and that the maximum \(\ell_2\)-norm distance between two probability vectors is $\sqrt{2}$.

Moreover,
\begin{align}
&\|\nabla_{\wb_\ic^{\ti}}\gc_\ic(\wb_\ic^{\ti};\siglb_\ic^{\ti})\|=\|\frac{2\lambda}{|\mathcal{P}|}\sum_{\xb\in\mathcal{P}}\nabla s(\wb_\ic^{\ti}, \xb)^T( s(\wb_\ic^{\ti}, \xb)-\siglb_\ic^{(t)}(\xb))+\nabla \lf(\wb_\ic^{\ti})\| \\
&\leq \frac{2\lambda}{|\mathcal{P}|}\|\sum_{\xb\in\mathcal{P}}\nabla s(\wb_\ic^{\ti}, \xb)^T( s(\wb_\ic^{\ti}, \xb)-\siglb_\ic^{(t)}(\xb))\|+\|\nabla \lf(\wb_\ic^{\ti})\|
\\
&\leq L_f+{2\lambda}\sum_{\xb\in\mathcal{P}}\|\nabla s(\wb_\ic^{\ti}, \xb)^T( s(\wb_\ic^{\ti}, \xb)-\siglb_\ic^{(t)}(\xb))\|\\
&\leq L_f+{2\lambda}\sum_{\xb\in\mathcal{P}}\|\nabla s(\wb_\ic^{\ti}, \xb)\|\|( s(\wb_\ic^{\ti}, \xb)-\siglb_\ic^{(t)}(\xb))\| \\ 
&\leq
L_f+{2\sqrt{2}\lambda}L_s|\mathcal{P}|=M_1
\end{align}
and therefore $\left\|\nabla_{\wb_\ic^{\ti}}\gc_\ic(\wb_\ic^{\ti};\siglb_\ic^{\ti})\right\|$ is bounded by $M_1\geq0$. With similar steps we can show that $\|\sgrad(\wb_\ic^{\ti};\siglb_\ic^{\ti})\|\leq M_2$ for a certain constant $M_2\geq0$. 
\end{proof}
\end{lemma}

\subsection{Main Proof for \Cref{the1}}
Using \Cref{lem1} and \Cref{lem2} we have that
\begin{flalign}
    &\expt[\gc_\ic(\wb_\ic^{\tip};\siglb_\ic^{\ti})|\mathcal{H}_t]=\expt[\gc_\ic(\wb_\ic^{\ti}-\lr \gb_\ic(\wb_\ic^{\ti};\siglb_\ic^{\ti}));\siglb_\ic^{\ti})|\mathcal{H}_t]\\
    &\leq \gc_\ic(\wb_\ic^{\ti};\siglb_\ic^{\ti})+\frac{\lr^2L_p}{2}\expt[\|\sgrad(\wb_\ic^{\ti};\siglb_\ic^{\ti})\|^2|\mathcal{H}_t]-\lr\nabla_{\wb_\ic^{\ti}}\gc_\ic(\wb_\ic^{\ti};\siglb_\ic^{\ti})^T\expt[\sgrad(\wb_\ic^{\ti};\siglb_\ic^{\ti})|\mathcal{H}_t]\\
&\leq \gc_\ic(\wb_\ic^{\ti};\siglb_\ic^{\ti})-\lr\|\nabla_{\wb_\ic^{\ti}}\gc_\ic(\wb_\ic^{\ti};\siglb_\ic^{\ti})\|^2
    +\frac{\lr^2L_p}{2}(2G^2+16\lambda^2L_s^2) \label{eq0-0}
\end{flalign}

Assuming $\sum_{t=0}^\infty\eta_t^2<\infty$ and $\sum_{t=0}^\infty\eta_t=\infty$, and applying Robbins-Siegmund Theorem~(Theorem B.1. in \cite{to2019stabi}) on \cref{eq0-0}, we have that with probability 1,
\begin{align}
    \sum_{t=1}^{\infty}\lr\|\nabla_{\wb_\ic^{\ti}}\gc_\ic(\wb_\ic^{\ti};\siglb_\ic^{\ti})\|^2<\infty
\end{align}
Now we can show
\begin{align}
    &\|\nabla\gc_\ic(\wb_\ic^{\tip};\siglb_\ic^{\ti})\|^2-\|\nabla\gc_\ic(\wb_\ic^{\ti};\siglb_\ic^{\ti})\|^2\\
    &=(\|\nabla\gc_\ic(\wb_\ic^{\tip};\siglb_\ic^{\ti})\|+\|\nabla\gc_\ic(\wb_\ic^{\ti};\siglb_\ic^{\ti})\|)(\|\nabla\gc_\ic(\wb_\ic^{\tip};\siglb_\ic^{\ti})\|-\|\nabla\gc_\ic(\wb_\ic^{\ti};\siglb_\ic^{\ti})\|)\\
    &\leq 2M_1(\|\nabla\gc_\ic(\wb_\ic^{\tip};\siglb_\ic^{\ti})\|-\|\nabla\gc_\ic(\wb_\ic^{\ti};\siglb_\ic^{\ti})\|)\\
    &\leq 2M_1\|\nabla\gc_\ic(\wb_\ic^{\tip};\siglb_\ic^{\ti})-\nabla\gc_\ic(\wb_\ic^{\ti};\siglb_\ic^{\ti})\|
    \\&\leq2M_1\|\lr\sgrad(\wb_\ic^{\ti};\siglb_\ic^{\ti})\|
    \leq 2M_1M_2\lr
\end{align}
Finally, using Proposition 2 in \cite{alb1998mp} we have that for $t\rightarrow\infty$,  $\|\nabla_{\wb_\ic^{\ti}}\gc_\ic(\wb_\ic^{\ti};\siglb_\ic^{\ti})\|\rightarrow0$ with probability 1.

\section{Proof for \Cref{the2}}
\label{app:the2proof}
We have that \cref{eq:lr} is equal to:
\begin{align}
    \wbl_\ic=\frac{1}{1+\lambda\nu/\beta}\wbt_\ic+\frac{1}{1+\beta/\lambda\nu}\sum_{i=1}^\ec\sw_{\ic,i}\wbt_i \label{eq22-1}
\end{align}
and the Bayes optimal $\wb_\ic$ in \cref{eq22-0} becomes
\begin{align}
    \wb_\ic=\frac{1}{1+\sigma^2/\beta\upsilon_\ic^2}\wbt_\ic+\frac{A_\ic\sigma^2}{\sigma^2+A_\ic\beta\upsilon_\ic^2}\sum_{i=1}^\ec\frac{1}{\sigma^2+\beta\upsilon_i^2}\wbt_i+\varsigma_\ic \label{eq22-2}
\end{align}
where $A_\ic=\left(\sum_{i\in[K],i\neq\ic}\frac{1}{\sigma^2+\beta\upsilon_i^2}\right)^{-1}$ and $\varsigma_\ic\sim\mathcal{N}\left(0,(\frac{\beta}{A_\ic+\beta\upsilon_\ic^2}+\frac{\beta}{\sigma^2})^{-1}\right)$. If we aim to find the $\lambda_\ic$ and $\sw_{\ic,i},i\in[K]$ that minimizes $\expt[F_\ic(\wbl_\ic)]$ given $\wbt_\ic$ and $\overline{\theta}_{\setminus \ic}$, in other words,
\begin{align}
    \lambda_\ic^*,\sw_{\ic,i}^*,i\in[K]&=\argmin_{\lambda_\ic,\sw_{\ic,i},i\in[K]}\expt[F_\ic(\wbl_\ic)|\wbt_\ic,\overline{\theta}_{\setminus \ic}]\\
    &=\argmin_{\lambda_\ic,\sw_{\ic,i},i\in[K]}\expt[\|\xxb_\ic\wbl_\ic-(\xxb_\ic\wb_\ic+\zb)\|_2^2|\wbt_\ic,\overline{\theta}_{\setminus \ic}]\\
    &=\argmin_{\lambda_\ic,\sw_{\ic,i},i\in[K]}\expt[\|\xxb_\ic(\wbl_\ic-\wb_\ic)\|_2^2|\wbt_\ic,\overline{\theta}_{\setminus \ic}]\\
    &=\argmin_{\lambda_\ic,\sw_{\ic,i},i\in[K]}\expt[\|\wbl_\ic-\wb_\ic\|_2^2|\wbt_\ic,\overline{\theta}_{\setminus \ic}] \label{eq22-3}
\end{align}
then taking \cref{eq22-1} and \cref{eq22-2} into \cref{eq22-3} we have that
\begin{align}
    \lambda_\ic^*=\sigma^2/\upsilon_\ic^2\nu\\
    \sw_{\ic,i}^*=\frac{B_k}{\sigma^2+\beta\upsilon_i^2}
\end{align}
where $B_k=\frac{A_\ic(\sigma^2+\beta\upsilon_\ic^2)}{\sigma^2+A_\ic\beta\upsilon_\ic^2}$.

\section{Further discussion on \Cref{the2}}
\label{app:the2dis}
\Cref{the2} presents insights on how to set the weights $\{\sw_{k,i}\}_{i\in[\ec]}$ and regularization weight $\lambda_\ic^*$ for each client $k\in[\ec]$ from \cref{eqn:main_objective} to improve generalization with \pktc. While in the main paper we discussed the implications of the optimal $\{\sw_{k,i}^*\}_{i\in[\ec]}$ in \cref{eqn:lropt} and the motivation for clustering, here we continue the discussion in regards to the optimal $\{\lambda_\ic^*\}_{\ic\in[\ec]}$, i.e., the optimal regularization weight. Recapping the linear regression setup from \Cref{sec:theoclust}, we have $\theta$ uniformly distributed on $\mathbb{R}^d$, and each device $\ic\in[\ec]$ has its data distributed with parameters $\wb_\ic=\theta+\zeta_\ic$ where $\zeta_\ic\sim\mathcal{N}(0,\upsilon_k^2\ib_d)$ and $\mathbf{I}_d$ is the $d\times d$ identity matrix and $\upsilon_\ic$ is unique to the client's task. Suppose we have $\yb_\ic=\xxb_\ic\wb_\ic+\zb,~\ic\in[K]$ where $\yb_\ic\in\mathbb{R}^n,~\xxb_\ic\in\mathbb{R}^{n\times d}$, and $\zb\in\mathbb{R}^n$ such that $\zb\sim\mathcal{N}(0,\sigma^2\ib_d)$. 

With \pktc, the optimal regularization weight is equal to $\lambda_\ic^*=\sigma^2/\upsilon_\ic^2\nu$ as shown in \Cref{the2}, where $\sigma^2$ and $\nu$ are constant across clients. This shows that clients with large $\upsilon_\ic$ can improve its generalization performance by having a smaller regularization weight. Intuitively, clients that have larger $\upsilon_\ic$ have a higher chance to have larger discrepancy in data distribution from other clients, and therefore having a smaller $\lambda_\ic$ can prevent from assimilating irrelevant knowledge from the other clients. Similarly, the opposite also holds where clients with smaller $\upsilon_\ic$ have a higher optimal $\lambda_\ic^*$. This result gives insight into how to set the regularization weight dependent on the client's data discrepancy to other clients. Although in our experiments we use identical $\lambda_\ic$ for $\ic\in[\ec]$, interesting future directions include varying $\lambda_\ic$ of across clients dependent on their data and training progress.

\section{Generalization Bound for Ensemble Models in Personalization} \label{app:genbound}

As defined in \Cref{sec:pkt}, we have the true data distribution of client $\ic$ defined as $\dd_\ic$, and the empirical data distribution associated with the client`s training dataset defined as $\ddh_\ic$. For a multi-class classification problem with a finite set of classes, we have that the data's domain is defined by the input space $\xb\in\mathcal{X}$ and the output space $y\in\mathcal{Y}$. For the generalization bound analysis we consider hypotheses that maps $h:\mathcal{X}\rightarrow\mathcal{Y}$, and $\mathcal{H}$ is defined as the hypotheses space such that $h\in\mathcal{H}$. The loss function $l(h(\xb),y)$ measures the classification performance of $h$ for a single data point $(\xb,y)$ and we define the expected loss over all data points that follow distribution $\mathcal{D}$ as $\mathcal{L}_\mathcal{D}(h)=\expt_{(\xb,y)\sim\mathcal{D}}[l(h(\xb),y)]$. We assume that $\gl(\cdot)$ is convex, and is in the range $[0,1]$. We define the minimizer of the expected loss over the data that follows the distribution $\dd_\ic$ and $\ddh_\ic$ as each $h_\ic=\argmin_{h}\mathcal{L}_{\dd_\ic}(h)$ and $\widehat{h}_\ic=\argmin\mathcal{L}_{\ddh_\ic}(h)$. Note that for sufficiently large training dataset, we will have $h_\ic\simeq\widehat{h}_\ic$. 

Our goal is to show the generalization bound for client $\ic$ such that $\gl_{\dd_\ic}\left(\sum_{i=1}^K\sw_{\ic,i}h_{\ddh_i}\right)$, where $h_{\ddh_i}$ represents the hypothesis trained from client $i$'s training dataset and $\sw_{\ic,i}$ represents the weight for the hypothesis of client $i$ for client $k$. For client $i\in[\ec]$, $h_{\ddh_i}$ will be the optimal hypothesis with respect to the training dataset for each client participating in FL, and the generalization bound for $\gl_{\dd_\ic}\left(\sum_{i=1}^K\sw_{\ic,i}h_{\ddh_i}\right)$ will show how the weighted average of different hypothesis from the other clients with respect to $\sw_{\ic,i},~i\in[K]$ helps the generalization of an individual client $\ic$ with respect to its true data distribution. Before presenting the generalization bound, we present several useful lemmas. 

\begin{lemma}[\textbf{Domain adaptation} \cite{dav2009domain}]
With two true distributions $\mathcal{D}_A$ and $\mathcal{D}_B$, for $\forall~\delta\in(0,1)$ and hypothesis $\forall h\in\mathcal{H}$, with probability at least $1-\delta$ over the choice of samples, there exists:
\begin{align}
    \gl_{\mathcal{D}_A}(h)\leq\gl_{\mathcal{D}_B}(h)+\frac{1}{2}d(\mathcal{D}_A,\mathcal{D}_B)+\nu
\end{align}
where $d(\mathcal{D}_A,\mathcal{D}_B)$ measures the distribution discrepancy between two distributions~\cite{dav2009domain} and $\nu=\inf_{h} \gl_{\mathcal{D}_A}(h)+\gl_{\mathcal{D}_B}(h)$.\label{lem3}
\end{lemma}

\begin{lemma}[\textbf{Generalization with limited training samples}]
For $\forall~\ic\in[\ec]$, with probability at least $1-\delta$ over the choice of samples, there exists:
\begin{align}
    \gl_{\dd_\ic}(h_{\ddh_\ic})\leq \gl_{\ddh_\ic}(h_{\ddh_\ic})+\sqrt{\frac{\log{2/\delta}}{2\ldat}}
\end{align}
where $\ldat$ is the number of training samples of client $\ic$. This lemma shows that for small number of training samples, i.e., small $\ldat$, the generalization error increases due to the discrepancy between $\dd_\ic$ and $\ddh_\ic$.
\begin{proof}
We seek to bound the gap between \(\mathcal{L}_{\mathcal{D}_k}(h_{\hat{\mathcal{D}}_k})\) and \(\mathcal{L}_{\hat{\mathcal{D}}_k}(h_{\hat{\mathcal{D}}_k})\). Observe that \(\mathcal{L}_{\mathcal{D}_k}(h_{\hat{\mathcal{D}}_k}) = \mathbb{E}\left[\mathcal{L}_{\hat{\mathcal{D}}_k}(h_{\hat{\mathcal{D}}_k})\right]\), where the expectation is taken over the randomness in the sample draw that generates \(\hat{\mathcal{D}}_k\), and that \(\mathcal{L}_{\hat{\mathcal{D}}_k}(h_{\hat{\mathcal{D}}_k})\) is an empirical mean over losses \(l(h(x), y)\) that lie within \([0,1]\). Since we are simply bounding the difference between a sample average of bounded random variables and its expected value, we can directly apply Hoeffding's inequality to obtain
\begin{align}
    \mathbb{P}\left[\mathcal{L}_{\hat{\mathcal{D}}_k}(h_{\hat{\mathcal{D}}_k}) - \mathcal{L}_{\mathcal{D}_k}(h_{\hat{\mathcal{D}}_k})  \geq \epsilon \right] \leq 2e^{-2m\epsilon^2}.
\end{align}
Setting the right hand side to \(\delta\) and rearranging gives the desired bound with probability at least \(1 - \delta\) over the choice of samples:
\begin{align*}
    \mathcal{L}_{\mathcal{D}_k}(h_{\hat{\mathcal{D}}_k}) & \leq \mathcal{L}_{\hat{\mathcal{D}_k}}(h_{\hat{\mathcal{D}}_k}) + \sqrt{\frac{\log 2/\delta}{2m_k}}.
\end{align*}
\end{proof} \label{lem4}
\end{lemma}
We now present the generalization bound for $\gl_{\dd_\ic}\left(\sum_{i=1}^\ec\sw_{\ic,i}h_{\ddh_i}\right)$ as follows:
\begin{align}
    \gl_{\dd_\ic}\left(\sum_{i=1}^\ec\sw_{\ic,i}h_{\ddh_i}\right)\leqt_{(c)}\sum_{i=1}^\ec\sw_{\ic,i}\gl_{\dd_\ic}(h_{\ddh_i})\leqt_{(d)}\sum_{i=1}^\ec\sw_{\ic,i}[\gl_{\dd_i}(h_{\ddh_i})+\frac{1}{2}d(\dd_i,\dd_\ic)+\nu_i] \label{eq1-0}
\end{align}
where $\nu_i=\inf_{h} \gl_{\mathcal{D}_i}(h)+\gl_{\mathcal{D}_\ic}(h)$, (c) is due to the convexity of $\gl$, and (d) is due to \cref{lem3}. We can further bound \cref{eq1-0} using \cref{lem4} as
\begin{align}
&\gl_{\dd_\ic}\left(\sum_{i=1}^\ec\sw_{\ic,i}h_{\ddh_i}\right)\leq\sum_{i=1}^\ec\sw_{\ic,i}\gl_{\ddh_i}(h_{\ddh_i})+\sum_{i=1}^\ec\sw_{\ic,i}\sqrt{\frac{\log{2/\delta}}{2\ldat}}+\frac{1}{2}\sum_{i=1}^\ec\sw_{\ic,i}d(\dd_i,\dd_\ic)+\sum_{i=1}^\ec\sw_{\ic,i}\nu_i
\\
&=\sum_{i=1}^\ec\sw_{\ic,i}\gl_{\ddh_i}(h_{\ddh_i})+\sqrt{\log{\delta^{-1}}}\sum_{i=1}^\ec\frac{\sw_{\ic,i}}{\sqrt{\ldat}}+\frac{1}{2}\sum_{i=1}^\ec\sw_{\ic,i}d(\dd_i,\dd_\ic)+\sum_{i=1}^\ec\sw_{\ic,i}\nu_i \label{eq1-1}
\end{align}
From \cref{eq1-1}, with $\gl_{\ddh_i}(h_{\ddh_i})$ in general being small for $\forall i\in[\ec]$ as it is the minimum loss, and $m_i$ being similar to other $m_{i'},~i'\in[\ec]$, the only way to minimize the generalization error of $\gl_{\dd_\ic}\left(\sum_{i=1}^\ec\sw_{\ic,i}h_{\ddh_i}\right)$ is to set the weights $\sw_{\ic,i}$ so that the third term $\frac{1}{2}\sum_{i=1}^\ec\sw_{\ic,i}d(\dd_i,\dd_\ic)$ is minimized. Note that it is difficult to know the value of $\nu_i$, making it impractical to minimize the fourth term in practice. This generalization results strengthens our motivation to use to find the weights $\sw_{\ic,i},~i\in[K]$ that minimizes $\frac{1}{|\mathcal{P}|}\sum_{\xb\in\mathcal{P}}\|\sum_{i=1}^\ec\sw_{\ic,i} s_i(\wb_i,\xb)-\sigl(\wb_\ic,\xb)\|_2^2$ in regards to the objective function we have in \cref{eqn:main_objective}.

\comment{
\subsection{Convergence Analysis for $\lf(\wb_\ic^{\ti})$}
We present the convergence of client $\ic$ for any $\ic\in[\ec]$ with respect to its own local objective function $\lf(\wb_\ic)$ as follows:

From \cref{as1}, we have that 
\begin{align}
    &\begin{aligned}\expt[\lf(\wb_\ic^{\tip})-\lf(\wb_\ic^{\ti})|\mathcal{H}_t]\leq\expt\left[\inner{\nabla\lf(\wb_\ic^{\ti})}{\wb_\ic^{\tip}-\wb_\ic^{\ti}}|\mathcal{H}_t\right]\\+\frac{L_p}{2}\expt\left[\|\wb_\ic^{\tip}-\wb_\ic^{\ti}\|_2^2|\mathcal{H}_t\right]
    \end{aligned}\\
    &\begin{aligned}
=-\eta\inner{\nabla\lf(\wb_\ic^{\ti})}{\expt\left[\frac{1}{
    |\midat|}\sum_{\xi \in \midat}\nabla \lff(\wb_\ic^{\ti}, \xi)+\hgrad(\wb_\ic^{\ti};\siglb^{\ti})|\mathcal{H}_t\right]}\\+\frac{L_p\eta^2}{2}\expt\left[\left\|\frac{1}{
    |\midat|}\sum_{\xi \in \midat}\nabla \lff(\wb_\ic^{\ti}, \xi)+\hgrad(\wb_\ic^{\ti};\siglb^{\ti})\right\|_2^2|\mathcal{H}_t\right]
    \end{aligned}\\ 
&\begin{aligned}
=-\eta\inner{\nabla\lf(\wb_\ic^{\ti})}{\expt\left[\frac{1}{
    |\midat|}\sum_{\xi \in \midat}\nabla \lff(\wb_\ic^{\ti}, \xi)|\mathcal{H}_t\right]}-\eta\inner{\nabla\lf(\wb_\ic^{\ti})}{\expt\left[\hgrad(\wb_\ic^{\ti};\siglb^{\ti})|\mathcal{H}_t\right]}\\+\frac{L_p\eta^2}{2}\expt\left[\left\|\frac{1}{
    |\midat|}\sum_{\xi \in \midat}\nabla \lff(\wb_\ic^{\ti}, \xi)\right\|_2^2|\mathcal{H}_t\right]+\frac{L_p\eta^2}{2}\expt\left[\left\|\hgrad(\wb_\ic^{\ti};\siglb^{\ti})\right\|_2^2|\mathcal{H}_t\right]\\+L_p\eta^2\expt\left[\inner{\frac{1}{
    |\midat|}\sum_{\xi \in \midat}\nabla \lff(\wb_\ic^{\ti}, \xi)}{\hgrad(\wb_\ic^{\ti};\siglb^{\ti})}|\mathcal{H}_t\right] 
\end{aligned}\\
&\begin{aligned}
\leq-\eta\left\|\nabla\lf(\wb_\ic^{\ti})\right\|_2^2-\eta(1-\eta L_p)\inner{\nabla\lf(\wb_\ic^{\ti})}{\expt\left[\hgrad(\wb_\ic^{\ti};\siglb^{\ti})|\mathcal{H}_t\right]}+\frac{L_p\eta^2G^2}{2}\\+\frac{\eta^2L_p}{2}\expt\left[\left\|\hgrad(\wb_\ic^{\ti};\siglb^{\ti})\right\|_2^2|\mathcal{H}_t\right] \label{eq0-0}
\end{aligned}
\end{align}
where \cref{eq0-0} uses \cref{as4}. With $\eta\leq\frac{1}{2L_p}$, we have from \cref{eq0-0} that
\begin{align}
\begin{aligned}
    \expt[\lf(\wb_\ic^{\tip})|\mathcal{H}_t]-\lf(\wb_\ic^{\ti})\leq-\eta\left\|\nabla\lf(\wb_\ic^{\ti})\right\|_2^2-\frac{\eta}{2}\inner{\nabla\lf(\wb_\ic^{\ti})}{\expt\left[\hgrad(\wb_\ic^{\ti};\siglb^{\ti})|\mathcal{H}_t\right]}\\+\frac{\eta^2L_p}{2}\expt[\|\hgrad(\wb_\ic^{\ti};\siglb^{\ti})\|_2^2|\mathcal{H}_t]+\frac{L_p\eta^2G^2}{2} \label{eq0-1}
    \end{aligned}\\
    \begin{aligned}
    \leq-\eta\left\|\nabla\lf(\wb_\ic^{\ti})\right\|_2^2+\frac{\eta}{4}\left\|\nabla\lf(\wb_\ic^{\ti})\right\|_2^2+\frac{\eta}{4}{\expt\left[\left\|\hgrad(\wb_\ic^{\ti};\siglb^{\ti})|\mathcal{H}_t\right\|_2^2\right]}\\+\frac{\eta^2L_p}{2}\expt[\|\hgrad(\wb_\ic^{\ti};\siglb^{\ti})\|_2^2|\mathcal{H}_t]+\frac{L_p\eta^2G^2}{2} \label{eq0-2}
    \end{aligned}\\
    \begin{aligned}
    \leq-\frac{3\eta}{4}\left\|\nabla\lf(\wb_\ic^{\ti})\right\|_2^2+\frac{\eta}{2}{\expt\left[\left\|\hgrad(\wb_\ic^{\ti};\siglb^{\ti})|\mathcal{H}_t\right\|_2^2\right]}+\frac{L_p\eta^2G^2}{2} \label{eq0-3}
    \end{aligned}
\end{align}
where \cref{eq0-2} uses Cauchy–Schwarz inequality and AM-GM inequality and \cref{eq0-3} uses $\eta\leq\frac{1}{2L_p}$.
Now we aim to bound the term $\expt[\|\hgrad(\wb_\ic^{\ti};\siglb^{\ti})\|_2^2|\mathcal{H}_t]$ in \cref{eq0-3}. We expand the term as
\begin{align}
    \|\hgrad(\wb_\ic^{\ti};\siglb^{\ti})\|_2^2=\left\|\frac{2\lambda}{|\mathcal{P}_\ic^{\ti}|}\sum_{x\in\mathcal{P}_\ic^{\ti}}\nabla s(\wb_\ic^{\ti}, x)^T\left( s(\wb_\ic^{\ti}, x)-\siglb^{(t)}(x)\right)\right\|_2^2\\
    \leq \frac{4\lambda^2}{|\mathcal{P}_\ic^{\ti}|}\sum_{x\in\mathcal{P}_\ic^{\ti}}\left\|\nabla s(\wb_\ic^{\ti}, x)^T\left( s(\wb_\ic^{\ti}, x)-\siglb^{(t)}(x)\right)\right\|_2^2 \label{eq0-4}\\
    \leq \frac{4\lambda^2}{|\mathcal{P}_\ic^{\ti}|}\sum_{x\in\mathcal{P}_\ic^{\ti}}\left\|\nabla s(\wb_\ic^{\ti}, x)\right\|_2^2\left\|s(\wb_\ic^{\ti}, x)-\siglb^{(t)}(x)\right\|_2^ 2 \label{eq0-5}\\
    \leq \frac{4\lambda^2L_s^2}{|\mathcal{P}_\ic^{\ti}|}\sum_{x\in\mathcal{P}_\ic^{\ti}}\left\|s(\wb_\ic^{\ti}, x)-\siglb^{(t)}(x)\right\|_2^2 \label{eq0-6}
\end{align}
where \cref{eq0-4} is due to Jensen's Inequality, \cref{eq0-5} is due to the submultiplicativy of the norm, and \cref{eq0-6} is due to \cref{as5}. With taking the expectation we have that

\begin{align}
\expt[\|\hgrad(\wb_\ic^{\ti};\siglb^{\ti})\|_2^2|\mathcal{H}_t]\leq \frac{4\lambda^2L_s^2}{|\mathcal{P}|}\sum_{x\in\mathcal{P}}\left\|s(\wb_\ic^{\ti}, x)-\siglb^{(t)}(x)\right\|_2^2={4\lambda^2L_s^2\rho(\wb_\ic^{\ti};\siglb^{(t)})} \label{eq0-7}
\end{align}
where $\rho(\wb_\ic^{\ti};\siglb^{(t)})=\frac{1}{|\mathcal{P}|}\sum_{x\in\mathcal{P}}\left\|s(\wb_\ic^{\ti}, x)-\siglb^{(t)}(x)\right\|_2^2$.
Plugging \cref{eq0-7} back to \cref{eq0-3}, we have that
\begin{align}
 \expt[\lf(\wb_\ic^{\tip})|\mathcal{H}_t]-\lf(\wb_\ic^{\ti})\leq-\frac{3\eta}{4}\left\|\nabla\lf(\wb_\ic^{\ti})\right\|_2^2+{2\eta\lambda^2L_s^2\rho(\wb_\ic^{\ti};\siglb^{(t)})}+\frac{L_p\eta^2G^2}{2}\label{eq0-8}\\
 \left\|\nabla\lf(\wb_\ic^{\ti})\right\|_2^2\leq\frac{4(\lf(\wb_\ic^{\ti})-\expt[\lf(\wb_\ic^{\tip})|\mathcal{H}_t])}{3\eta}+{2\eta\lambda^2L_s^2\rho(\wb_\ic^{\ti};\siglb^{(t)})}+\frac{L_p\eta^2G^2}{2}\label{eq0-9}
\end{align}
Summing both sides of \cref{eq0-9} for $i=1,...,t$, with \cref{as2} and rearranging we have
\begin{align}
    \frac{1}{t}\sum_{i=1}^{t}\|\nabla\lf(\wb_\ic^{(i)})\|_2^2\leq\frac{4(\lf(\wb_\ic^{(1)})-F_{\ic,\inf})}{3\eta t}+\frac{2L_p\eta G^2}{3}+\frac{8\lambda^2L_s^2}{3t}\sum_{i=1}^{t}\rho(\wb_\ic^{(i)};\siglb^{(i)}) \label{eq0-9-1}
\end{align}

}

\section{Details of Experimental Setup } \label{app:exp}
Codes for the results in the paper are presented in the supplementary material. 

\subsection{Description for Toy Example - \Cref{fig:lrtoy}}
For \Cref{fig:lrtoy}, we design a linear regression problem where the true local model for each client is generated as $\wb_i=\theta+\zeta_i,i\in[3]$ where $\theta\in\mathbb{R}^{2\times 1}$ is a non-informative prior which elements are uniformly distributed $\mathcal{U}(-10,10)$ and the elements of $\zeta_i$ follows the normal distribution $\mathcal{N}(0,\sigma_i),~\sigma_1=2,\sigma_2=5,\sigma_3=200$. The discrepancy across the variance denotes the data-heterogeneity across the clients. The range for $\xb$ is $[-10,10]$ for all elements. We assume all clients have identical dataset size. For implementing \pkt for the toy example, we set the public data range as identical to the input data range, and set $\lambda=50$. For KT w/o clustering, the co-distillation term uses a simple average of all the logits from the clients for regularizing while for KT with clustering the weights are set so that clients with similar true local models have higher weights for each other. This setting is also consistent with the generalization analysis presented in \Cref{sec:theoclust}. The code for the linear regression toy example is presented in the supplementary material.

\subsection{Description for CIFAR10 Experiments.}

\paragraph{Data Partitioning.} We experiment with three different seeds for the randomness in the dataset partition across clients and present the averaged results across the seeds with the standard deviation. The partitioning of each individual client's data to training/validation/test dataset is done as follows: after partitioning the entire dataset by the Dirichlet distribution $\text{Dir}_{K}(\alpha)$ with $\alpha=0.01$ across clients, we partition each client dataset by a $\{0.1,0.3,0.4\}/0.1/0.5$ ratio where the ratio for the training dataset is chosen by random from $\{0.1,0.3,0.4\}$ for each client. Such partitioning simulates a more realistic FL setting where individual clients may not have sufficient labeled data samples for training that represents the test dataset's distribution. For all experiments we run 500 communication rounds which we have observed allows convergence for all experiments. 

\paragraph{Local Training and Hyperparameters.} For the local-training hyperparameters, we do a grid search over the learning rate: $\eta\in\{0.1, 0.05, 0.01, 0.005, 0.001\}$, batchsize: $b\in\{32, 64, 128\}$, and local iterations: $\tau\in\{10, 30, 50\}$ to find the hyper-parameters with the highest test accuracy for each benchmark. For all benchmarks we use the best hyper-parameter for each benchmark after doing a grid search over feasible parameters referring to their source codes that are open-sourced. For the knowledge distillation server-side hyperparameters, we do a grid search over the public batch size: $b'\in\{32, 64, 128\}$, regularization weight $\lambda\in\{0.05, 0.1, 0.5, 1, 2, 4\}$ to find the best working hyperparameters. The best hyperparameters for \pkt we use is $\eta=0.001,b=64,\tau=50,b'=128,\lambda=2$. 

\paragraph{Model Setup.} For the model configuration, for the CNN we have a self-defined convolutional neural network with 2 convolutional layers with max pooling and 4 hidden fully connected linear layers of units $[120,100,84,50]$. The input is the flattened convolution output and the output is consisted of 10 units each of one of the 0-9 labels. For the VGG, we use the open-sourced VGG net from Pytorch with torchvision ver.0.4.1 presented in Pytorch without pretrained as False and batchnorm as True.

\paragraph{Platform.} All experiments are conducted with clusters equipped with one NVIDIA TitanX GPU. The number of clusters we use vary by $C$, the fraction of clients we select. The machines communicate amongst each other through Ethernet to transfer the model parameters and information necessary for client selection. Each machine is regarded as one client in the federated learning setting. The algorithms are implemented by PyTorch.


\end{document}